\documentclass[11pt]{article}
\usepackage{listings}
\usepackage{tabularx}

\usepackage[preprint]{acl}

\usepackage{times}
\usepackage{tikz}
\usepackage{latexsym}

\usepackage{amsmath}
\usepackage{hyperref}       % hyperlinks
\usepackage{url}            % simple URL typesetting
\usepackage{booktabs}       % professional-quality tables
\usepackage{amsfonts}       % blackboard math symbols
\usepackage{nicefrac}       % compact symbols for 1/2, etc.
\usepackage{microtype}      % microtypography
\usepackage{xcolor}         % colors
\usepackage{subfig}

\usepackage[most]{tcolorbox}
\usepackage{xcolor}

\usepackage{fvextra}

\definecolor{promptbg}{RGB}{248,249,250}
\definecolor{promptframe}{RGB}{80,120,180}
\definecolor{prompttitle}{RGB}{30,70,130}

\newtcolorbox{promptbox}[2][]{
    enhanced,
    colback=promptbg,
    colframe=promptframe,
    coltitle=white,
    colbacktitle=prompttitle,
    title={#2},
    fonttitle=\bfseries,
    boxrule=0.6pt,
    arc=2pt,
    left=6pt,
    right=6pt,
    top=6pt,
    bottom=6pt,
    before skip=6pt,
    after skip=6pt,
    #1
}

\usepackage{amssymb}
\usepackage{mathtools}
\usepackage{amsthm}
\theoremstyle{plain}

\theoremstyle{definition}

\theoremstyle{remark}

\usepackage[T1]{fontenc}
\usepackage[utf8]{inputenc}

\usepackage{microtype}

\usepackage{inconsolata}

\usepackage{graphicx}

\title{Many Voices, One Reward: Multi-Role Rubric Generation for LLM Judging and Reward Modeling}

\author{
  \textbf{Dazhi Fu\textsuperscript{1,2}},
  \textbf{Jiuding Yang\textsuperscript{2}},
  \textbf{Yiwen Guo\textsuperscript{3,*}},
  \textbf{Jicong Fan\textsuperscript{1,*}}
\\
  \textsuperscript{1}School of Data Science, The Chinese University of Hong Kong, Shenzhen, China\\
  \textsuperscript{2}LIGHTSPEED\\
  \textsuperscript{3}Independent Researcher\\
\\
   \small{
   \href{mailto:dazhifu@link.cuhk.edu.cn}{dazhifu@link.cuhk.edu.cn},
   \href{mailto:jiudingyang@tencent.com}{jiudingyang@tencent.com}
}\\
\small{
   \href{mailto:guoyiwen89@gmail.com}{guoyiwen89@gmail.com}, 
   \href{mailto:fanjicong@cuhk.edu.cn}{fanjicong@cuhk.edu.cn}
  }
}

\begin{document}
\maketitle
\begingroup
\renewcommand{\thefootnote}{\fnsymbol{footnote}}
\footnotetext[1]{indicates corresponding authors}
\endgroup

% Required packages:
% \usepackage{booktabs}
% \usepackage{xcolor}
% \usepackage{enumitem}
% \usepackage{array}

%Classical reward models \cite{ouyang2022training, frick2024evaluate} typically compress rich, multi-dimensional notions of response quality into a single scalar.  The LLM-as-a-judge paradigm \cite{zheng2023judging, dubois2023alpacafarm} leverages an LLM's broad world knowledge to score responses directly, but off-the-shelf judges suffer from position bias, verbosity bias, self-preference, and opaque decision criteria \cite{wang2024large,panickssery2024llm}. In both cases, the core issue is the same: the evaluation signal is implicit and entangled, making it hard to audit, calibrate, and optimize.

\begin{abstract}

Reliable reward and preference signals are critical for evaluating and optimizing large language models on open-ended tasks. Rubric-based judges offer a transparent way to decompose such judgments into explicit evaluation criteria, but existing annotation-free rubric generators typically rely on a single generic evaluator. As a result, they may overlook important dimensions of human preference, a failure mode we term \emph{dimensional blind spots}. To address this limitation, we propose Multi-Role Rubric Generation (MRRG), a training-free and reference-free framework that elicits evaluation criteria from multiple complementary roles and consolidates them into an auditable rubric-based scorer. This scorer can be used both to validate pairwise preferences and to provide rewards for GRPO-style Reinforcement Learning with Verifiable Rewards (RLVR). Experiments on preference validation benchmarks show that MRRG consistently outperforms single-role rubric generation baselines across multiple backbone models. Further RLVR experiments demonstrate that MRRG yields a stronger reward signal for improving open-ended generation. Our code is available \href{https://github.com/fudazhiaka/Many-Voices-One-Reward-Multi-Role-Rubric-Generation-for-LLM-Judging-and-Reward-Modeling}{here}.

\end{abstract}

\section{Introduction}
\label{sec:intro}
Large language models (LLMs) \cite{brown2020language,achiam2023gpt,yang2025qwen3} have achieved remarkable capabilities across a wide range of open-ended generation tasks. A key driver of this progress is post-training, which aligns pretrained models with human intentions through techniques such as Reinforcement Learning with Human Feedback (RLHF) \cite{ouyang2022training,bai2022training} and Reinforcement Learning with Verifiable Rewards (RLVR) \cite{lambert2024tulu,guo2025deepseek}. Despite their differences in reward acquisition, both approaches share a common dependency: reliable reward or preference signals that distinguish better responses from worse ones. Constructing such signals is especially challenging for open-ended tasks, where response quality is multidimensional, context-dependent, and difficult to reduce to a single opaque score. How these signals are defined, produced, and validated has therefore emerged as a central challenge for both evaluation \cite{zheng2023judging,lambert2025rewardbench} and optimization.

A natural way to make these signals more transparent — inspired by how humans evaluate open-ended work — is to express evaluation criteria as rubrics or checklists \cite{viswanathan2025checklists,que2024hellobench,saha2024branch}. Rather than collapsing quality into a single opaque judgment, a rubric decomposes evaluation into concrete, verifiable sub-questions. Prior work has shown that rubric- or criteria-conditioned judges can substantially narrow the gap with human raters \cite{shen2026rethinking,kawabata2026c2}, and that rubric-style process supervision can stabilize reasoning-oriented reinforcement learning \cite{lightman2023let,uesato2022solving}. Together, these results position rubrics as a promising interface for both evaluation and optimization. However, existing approaches typically rely on gold responses, expert-written references, large-scale preference datasets, or extensive human annotations to construct effective rubrics — requirements that are costly and frequently unavailable in realistic open-ended settings.

Recent annotation-free approaches \cite{zhang2025chasing,shen2026rethinking} reduce this cost by generating rubrics directly from the query. Yet they typically rely on a single generic evaluator to enumerate all quality criteria at once. This \emph{single-voiced} formulation assumes that one evaluator can recover the full structure of human preference, overlooking the fact that human judgment is inherently distributed across complementary perspectives: users emphasize usefulness, domain experts prioritize correctness, and educators focus on clarity, evidence, and safety. Consequently, single-voiced rubrics tend to omit preference-relevant dimensions — a failure mode we term \emph{dimensional blind spots}.

\begin{figure*}[t]
\centering
\subfloat[Domain-dependent judging accuracy in single-voiced rubric]{
    \includegraphics[scale=0.4]{"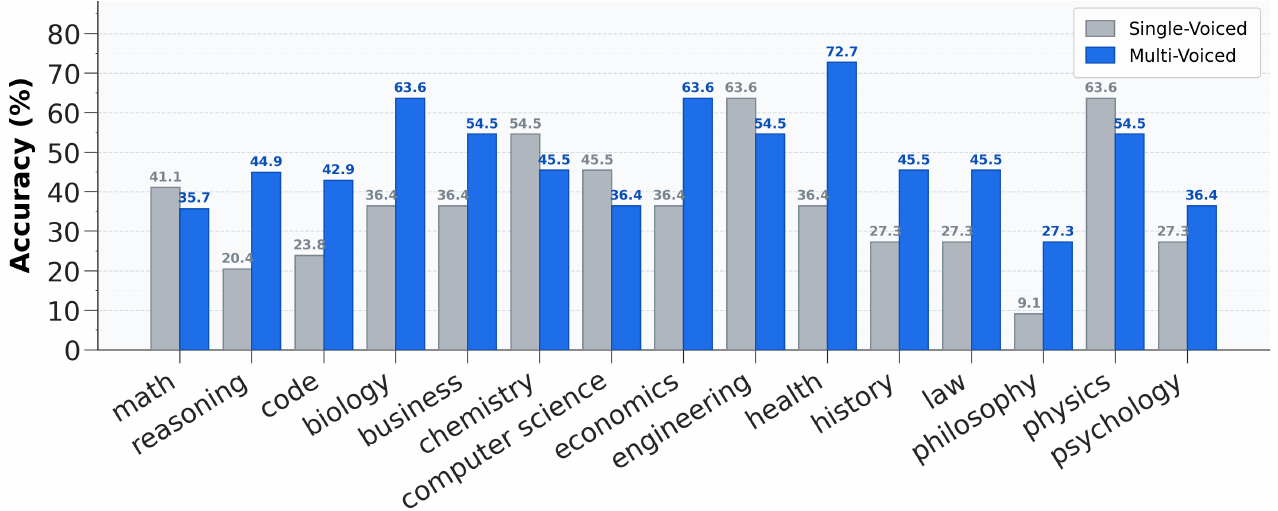"}
    \label{fig:sub_acc}
}
\hfill
\subfloat[Rubric hacking]{
    \includegraphics[scale=0.35]{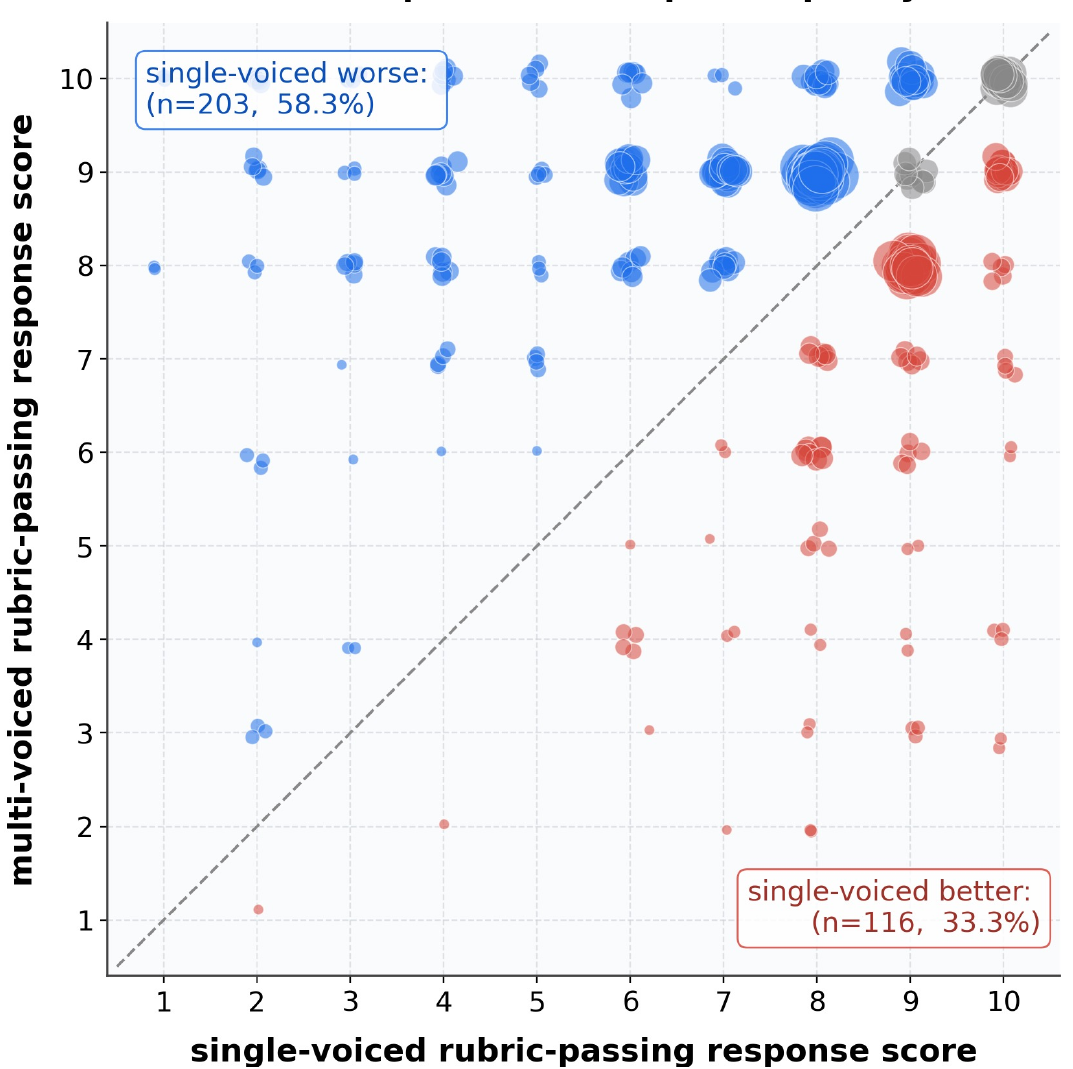}
    \label{fig:sub_rubric}
}
\caption{Two main problems caused by dimensional blind spots in single-voiced rubrics.}
\label{fig:problems}
%\vspace{-15pt}
\end{figure*}

Dimensional blind spots lead to two concrete failures. First, they make judging accuracy highly domain-dependent: as shown in Figure~\ref{fig:sub_acc}, single-voiced rubric judgment performs well in technically structured domains but degrades when domain-specific quality criteria are insufficiently covered. Second, they enable \emph{rubric hacking}: as shown in Figure~\ref{fig:sub_rubric}, responses can score well by optimizing for covered criteria while remaining low quality. Table~\ref{tab:nasdaq_answer_comparison} illustrates this mechanism concretely: for a broad NASDAQ-related question, a single-voiced rubric rewards surface-level task completion while overlooking feasibility, epistemic calibration, and unverifiable claims. Additional analyses are provided in Appendix~\ref{app-experiment}.

\begin{figure*}[t!]
    \centering
    \includegraphics[width=0.9\linewidth]{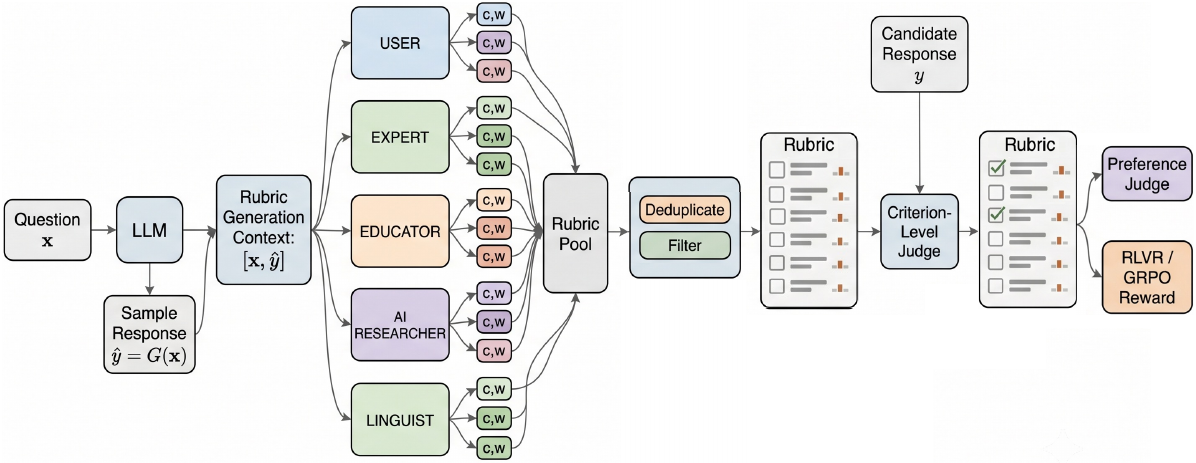}
    \caption{Framework of MRRG.}
    \label{figure:framework}
\end{figure*}

To address these blind spots, we propose \emph{Multi-Role Rubric Generation} (MRRG), a training-free and reference-free framework that recasts rubric generation as multi-role elicitation. Rather than relying on a single generic evaluator, MRRG instructs an LLM to adopt complementary evaluative roles — such as user, domain expert, educator, AI researcher, and linguist — each generating compact, atomic, and verifiable rubric items. These are then pooled and deduplicated into an auditable rubric-based scorer suitable for both preference validation and as a reward model in GRPO-style RLVR. Figure~\ref{figure:framework} illustrates the overall framework.
Our contributions are as follows:
\begin{itemize}
\item \textbf{Dimensional blind spots.} We identify dimensional blind spots as a key failure mode of single-voiced rubric generation and demonstrate that they cause domain-dependent judging accuracy and rubric hacking, as illustrated in Figure~\ref{fig:problems}.
\item \textbf{Multi-Role Rubric Generation.} We propose MRRG, a training-free and reference-free framework that elicits rubric criteria from multiple complementary roles and aggregates them into an auditable scorer for both LLM-as-a-judge evaluation and RLVR reward modeling.
\item \textbf{Empirical validation.} On RewardBench-2, JudgeBench, and PPE, MRRG consistently outperforms the strongest listed baseline across Qwen2.5 and GPT-series models, with gains of 3.1–16.4 percentage points. In GRPO-style RLVR experiments, MRRG improves over the strongest single-voice reward baseline by 1.7 points on BiGGen Bench and 3.4 points on HealthBench-Hard.
\end{itemize}

\begin{table*}[h!]
    \centering
    \scriptsize
    \renewcommand{\arraystretch}{1.05}
    \setlength{\tabcolsep}{5pt}
    \caption{Case study comparing single-voiced and multi-voiced rubrics with corresponding passing answers. \textbf{i} in \textbf{[Wi]} indicates the weight for each criteria.}
    \label{tab:nasdaq_answer_comparison}

    \begin{tabular}{p{0.48\textwidth}|p{0.48\textwidth}}
        \toprule
        \multicolumn{2}{c}{
        \textbf{Question:} Can you list and analyse the investment strategies and history of operation for each company listed on the NASDAQ?
        } \\
        \midrule

        \textbf{Single-voiced rubric} 
        & 
        \textbf{Multi-voiced rubric} \\
        \midrule

        \begin{itemize}
          \item  Lists all NASDAQ-listed companies. \textbf{[W3]}
          \item  Describes investment strategies for each company. \textbf{[W3]}
          \item  Provides historical analysis of operations for each company. \textbf{[W3]}
          \item  Uses reliable and verifiable sources. \textbf{[W2]}
          \item  Organizes information clearly, e.g., by company. \textbf{[W2]}
          \item  Avoids irrelevant or extraneous information. \textbf{[W2]}
          \item Uses precise and professional language.  \textbf{[W1]}
          \item  Includes recent and up-to-date information. \textbf{[W2]}
          \item  Clarifies technical terms or jargon. \textbf{[W1]}
        \end{itemize}
        &
        \begin{itemize}
          \item { Acknowledges impracticality of covering every company; sets scope boundaries. \textbf{[W3]}}
          \item { Avoids unsupported generalizations about companies. \textbf{[W2]}} 
          \item { Flags potentially outdated or incomplete data. \textbf{[W2]}}
          \item { Provides a methodology/framework instead of exhaustive listing. \textbf{[W3]}}
          \item { Avoids unverifiable or speculative claims. \textbf{[W3]}}
          \item { Clearly defines ``investment strategies'' and ``history of operation''. \textbf{[W3]}}
          \item { Avoids misleading implications or over generalizations. \textbf{[W3]}}
          \item { Avoids conflating market trends with individual strategies. \textbf{[W3]}}
          \item { Acknowledges if the request is too broad or infeasible. \textbf{[W3]}}
        \end{itemize}
        \\
        \midrule

        \textbf{Single-voiced rubric passing answer} 
        & 
        \textbf{Multi-voiced rubric passing answer} \\
        \midrule

        \textbf{Apple Inc.}: strategy centers on brand loyalty, ecosystem integration, and R\&D; history includes the Macintosh, iPod, iPhone, and iPad. 
        \textbf{Amazon.com, Inc.}: strategy spans e-commerce, AWS, logistics, and streaming; history includes its start as an online bookstore, AWS launch, and Whole Foods acquisition. 
        This gives a brief company-by-company overview.
        &
        A complete firm-by-firm analysis of all NASDAQ-listed companies is impractical in one response. A better answer defines the scope, selects firms or sectors, and analyzes each using filings, annual reports, investor presentations, and market data. Key dimensions include capital allocation, acquisitions, R\&D, revenue model, operational milestones, risks, and recent performance.
        \\
        \bottomrule
    \end{tabular}
\end{table*}

\section{Method}
\label{sec:method}

This section develops MRRG in four steps. We first formalize
\emph{rubric-based scoring} as a unified interface that subsumes both
LLM-as-a-judge and RLVR reward modeling (Sec.~\ref{sec:method:framework}).
We then introduce our multi-role rubric generator, MRRG
(Multi-Role Rubric Generation, Sec.~\ref{sec:method:mp-rubric}), and
describe how the resulting per-role rubrics are pooled into a
scalar score (Sec.~\ref{sec:method:aggregation}) that can be
used, without modification, as both a preference validator and an RLVR
reward (Sec.~\ref{sec:method:usage}). 

\subsection{Preliminaries: Rubric-Based Scoring}
\label{sec:method:framework}

Let $x\in\mathcal{X}$ denote an input prompt and $y\in\mathcal{Y}$ a
candidate response. Every alignment signal aims to approximate the latent
human quality function
$q^{\star}:\mathcal{X}\times\mathcal{Y} \to \mathbb{R}$. Classical reward
models parametrize $q^{\star}$ as a scalar network $r_\theta(x,y)$ trained
on preference pairs~\citep{ouyang2022training,bai2022training};
LLM-as-a-judge approaches instantiate it as a prompted language model,
$q^{\star}(x,y)\approx G\big({prompt}_{{judge}}(x,y)\big)$~\citep{zheng2023judging}.
Both collapse the multi-dimensional structure of $q^{\star}$ into an
opaque scalar.

\paragraph{Notation.}

Throughout the paper, $G(\cdot)$ denotes a call to a pretrained language
model. Different uses of $G$ are distinguished by the prompt template
provided to the model. We use ${prompt}_{{judge}}$ to denote
the per-criterion evaluation template, ${prompt}_{m}$ for
role-conditioned rubric generation under role $p_m$, and
${prompt}_{r}$ for generic rubric generation.

\paragraph{Rubric-based scoring.}
A rubric-based scorer breaks this collapse by introducing an
interpretable intermediate variable. A \emph{rubric}
$R(x)=\{c_{k},w_{k}\}_{k=1}^K$ is a set of \emph{criteria}, each $c_k$
being an atomic, verifiable proposition about the response (e.g.,
\emph{``The answer cites the correct theorem''}). Each criterion is
evaluated by the same language model to
$s_k=G\big({prompt}_{{judge}}(x,y,c_k)\big)\in\{0,1\}$, and the
overall score aggregates criterion-level judgments:
\begin{equation}
\label{eq:rubric-score}
\begin{aligned}
    &S(x,y;R)\;\\&=\;\frac{\sum_{k=1}^{K} w_{k}\cdot G\big({prompt}_{{judge}}(x,y,c_k)\big)}{\sum_{k=1}^{K} w_{k}},
\end{aligned}
\end{equation}
where $w_{k}>0$ is the importance of the $k$-th criterion. This
decomposition makes the signal simultaneously \emph{auditable} (every
contribution $w_k s_k$ is attributable to a named criterion),
\emph{verifier-like} (each $s_k$ is a constrained yes/no or graded
judgment), and \emph{model-agnostic} (any sufficiently capable LLM can
serve as $G$). The central design question thus becomes: \emph{how
should $R(x)$ be constructed so as to cover more comprehensive evaluative dimensions and thus be  close to $q^{\star}$?}

\subsection{Multi-Role Rubric Generation}
\label{sec:method:mp-rubric}

Existing rubric
generators~\citep{gunjal2025rubrics,zhang2025chasing,liu2025openrubrics}
construct initial $R(x)$ with one or a few LLM calls, e.g.,
$R(x)=G\big({prompt}_{r}\big(x,\hat{y}\big)\big)$ with
$\hat{y}=G(x)$ a sample response, where a ``universal evaluator'' is
asked to enumerate every relevant criterion at once. As argued
in Sec.~\ref{sec:intro}, this single-voice formulation compresses genuinely distinct viewpoints
into a shared prior and could lead to dimensional blind spots.

\paragraph{Role-conditioned elicitation.}
A straightforward approach to broadening the evaluative dimensions in rubric generation is to repeat the generation process multiple times. Although this can increase the apparent diversity of the resulting rubrics, it remains constrained by the inherent biases and limitations of a single evaluative perspective or role. To address this limitation, we recast rubric generation as a \emph{multi-role elicitation} problem.
Let $\mathcal{P}=\{p_1,\ldots,p_M\}$ be a small, fixed set of evaluative
\emph{roles}; by default
$\mathcal{P}=\{\textsc{User},\ \textsc{Domain Expert},\
\textsc{Educator},\\ \textsc{AI Researcher},\ \textsc{Linguist}\}$. For each
role $p_m$ we supply a role-specific template ${prompt}_m$ that
foregrounds the concerns most salient to that role---e.g., the
\textsc{User} template emphasizes problem-solving utility and intent
satisfaction, while the \textsc{Educator} template emphasizes evidential support, communication quality and safety. This produces a
role-specific rubric
\begin{equation}
\label{eq:per-role}
\begin{aligned}
    R_{m}(x)\;&=\;G\big({prompt}_m(x,\hat{y})\big)
\;\\&=\;\{(c_{m,k},w_{m,k})\}_{k=1}^{K_m},
\end{aligned}
\end{equation}
where $m=1,\ldots,M$ and $\hat{y}=G(x)$ is a sample response. Each $R_m(x)$ is compact
($K_m\in[3,7]$ in practice) and internally coherent, because the
generator only has to reason from one viewpoint at a time. 
%Full prompt templates are given in Appendix~\ref{app:prompts}. 
Importantly, $w_{m,k}$ is produced \emph{jointly} with $c_{m,k}$ inside the same
role-conditioned call: the role that writes a criterion is also the
one best positioned to judge its importance.

\paragraph{Criterion format and self-assessed importance.}
Each role is also instructed  to
emit \emph{binary} criteria paired with a \emph{self-assessed}
importance weight $w_{m,k}\in\{1,2,3\}$ on a three-point scale:
\begin{equation}
\label{eq:weight-scale}
w_{m,k}=
\begin{cases}
3 & \text{core need}\\
2 & \text{significant issue }\\
1 & \text{non-critical polish}
\end{cases}
\end{equation}

\paragraph{Selection of Roles.}

One might consider selecting a question-specific subset of roles, since
the perspectives most relevant to evaluation can in principle depend on
the input question. We instead adopt a fixed but carefully chosen role
pool $\mathcal{P}$, motivated by four considerations.
\emph{(i)} Determining the ``most relevant'' roles for an arbitrary
question is itself ill-posed and would likely require an LLM-based role
selector. Such a selector could inherit, or even amplify, the same
dimensional biases that MRRG aims to mitigate.
\emph{(ii)} The \textsc{Domain Expert} role is already instantiated
conditionally on the input question, providing question-aware
specialization without exposing the high-level role-selection process to
additional model bias.
\emph{(iii)} The default roles cover the principal axes along which humans
evaluate open-ended generation: \emph{practical utility}
(\textsc{User}), \emph{technical correctness} (\textsc{Domain Expert}),
\emph{communicative clarity} (\textsc{Educator}),
\emph{AI-specific reasoning and model behavior}
(\textsc{AI Researcher}), and \emph{linguistic quality}
(\textsc{Linguist}).
\emph{(iv)} Keeping the role pool small, with $|\mathcal{P}|=M$ and
$M=5$ in our experiments, bounds the rubric-generation cost by
$\mathcal{O}(M)$ LLM calls per prompt, independent of dataset size.
We study the sensitivity of MRRG to the choice of $\mathcal{P}$ in
Section~\ref{sec:ablation}.

%One might consider selecting a different role subset for each
%question, since the perspectives relevant to evaluation can in
%principle be question-dependent. We instead adopt a fixed but
%carefully chosen pool $\mathcal{P}$, motivated by four
%considerations.
%\emph{(i) } Determining the
%``most relevant'' roles for an arbitrary question is ill-posed and
%would have to rely on an LLM selector, which risks inheriting the
%very dimensional biases MRRG aims to remove.
%\emph{(ii)} The \textsc{Domain Expert} role is
%already instantiated conditionally on the input question, providing
%question-aware specialization without exposing the high-level
%perspective space to model bias.
%\emph{(iii)} The five default roles span the principal
%axes along which humans evaluate open-ended generation---\emph{practical
%utility} (\textsc{User}), \emph{technical correctness}
%(\textsc{Domain Expert}), \emph{communicative clarity}
%(\textsc{Educator}), 
%\emph{AI-specific reasoning and model behavior} (\textsc{AI Researcher}),
%and \emph{linguistic quality} (\textsc{Linguist}).
%\emph{(iv)} Fixing $|\mathcal{P}|=M$ small (we use
%$M{=}5$) caps rubric-generation cost at $\mathcal{O}(M)$ LLM calls
%per prompt, independent of dataset size.
%We ablate the sensitivity to $\mathcal{P}$ in Section~\ref{sec:ablation}.

\subsection{Rubric Deduplication}
\label{sec:method:aggregation}

The per-role rubrics $\{R_m(x)\}_{m=1}^{M}$ may contain repeated
criteria, since different roles can independently produce the same
requirement. In practice, we also observe that LLMs sometimes generate
identical criteria within  role-specific rubrics, potentially due
to hallucinated redundancy. If left untreated,
such exact duplicates may be counted multiple times and distort the final
rubric-based judgment. To address this issue, we apply a simple
deterministic post-processing step.

We first concatenate all role-specific rubrics into a single list,
\begin{equation}
\label{eq:concat_rubrics}
    R_{1:M}(x)
    =
    \big[
    (c_{m,k}, w_{m,k})
    \big]_{m=1,k=1}^{M,K_m}.
\end{equation}
We then scan this list in order and remove criteria whose text exactly
matches a criterion that has already appeared. The first occurrence is
kept together with its original weight, while subsequent duplicate
occurrences are discarded. The final rubric is
\begin{equation}
\label{eq:deduplicated_rubric}
    \widetilde{R}(x)
    =
    \big[
    (\Tilde{c}_{j}, \Tilde{w}_{j})
   \big]_{j=1}^{N},
\end{equation}
where $N$ is the number of criteria in the final rubric.

Importantly, this procedure only removes exact duplicates and leaves all
non-identical criteria unchanged, including criteria that are semantically
similar or potentially in tension with one another. Although one could
apply semantic deduplication or consolidation methods, such as using an
LLM to merge redundant criteria or resolve conflicting ones, we avoid this
design choice for two reasons. First, LLM-based consolidation introduces
an additional judgment step, where decisions about redundancy and conflict
may themselves be biased and may propagate such bias into the final
rubric. Second, merging or removing semantically similar criteria from
different roles may inadvertently weaken important evaluation dimensions:
when multiple roles independently emphasize similar criteria, this
agreement can signal that the criterion is important. Collapsing these
criteria into a single item may reduce their effective influence and thus
affect the final judgment. We therefore adopt exact-match deduplication as
a conservative, deterministic, and reproducible choice. Additional
empirical analysis is provided in Appendix \ref{sec:ablation_consolidation}.

\subsection{From Rubric Score to Alignment Signal}
\label{sec:method:usage}

The scalar $S(x,y;\widetilde{R}(x))\in[0,1]$ plugs into both downstream
settings without modification.

\paragraph{Preference validation (LLM-as-a-judge).}
Given two candidates $(y_A,y_B)$ for the same prompt $x$, we predict the
preferred response as
$\arg\max_{i\in\{A,B\}} S(x,y_{i};\widetilde{R}(x))$.

\paragraph{Reward modeling for RLVR.}
We use $S(x,y;\widetilde{R}(x))$ directly as the reward inside
GRPO-style optimization~\citep{guo2025deepseek}.
For a prompt $x$ with a rollout group $\{y_i\}_{i=1}^{G_{\text{roll}}}$,
the group-relative advantage of $y_i$ is
\begin{equation}
\label{eq:grpo-advantage}
A_{i}\;=\;\frac{S(x,y_{i};\widetilde{R}(x))-\mu_{S}(x)}
{\sigma_{S}(x)+\varepsilon},
\end{equation}
where $\mu_{S}(x)$ and $\sigma_{S}(x)$ denote the mean and standard
deviation of $\{S(x,y_{i};\widetilde{R}(x))\}_{i=1}^{G_{\text{roll}}}$
taken \emph{within the rollout group for prompt $x$} and $\varepsilon$
is a small constant for numerical stability. Because
$\widetilde{R}(x)$ is generated once per prompt and cached, the per-rollout
reward cost reduces to $N$ yes/no judge calls, which
parallelize trivially across the group.

\section{Experimental Results}
\label{sec:exp}

We evaluate MRRG along the two downstream settings it was
designed for: (i) \emph{preference validation}, where MRRG
acts as an LLM-as-a-judge on benchmark preference pairs
(Sec.~\ref{sec:exp:judge}); and (ii) \emph{reward modeling for RLVR},
where the same rubric-based scorer drives GRPO-style policy
optimization (Sec.~\ref{sec:exp:rlvr}). Across the experimental part we
answer the following questions:

\begin{itemize}
\item[\textbf{Q1.}] Can MRRG outperform single-voiced rubric
generators and other baselines in both preference validation and RLVR
reward modeling?
\item[\textbf{Q2.}] Is MRRG robust to the choice of roles, or does it rely
critically on any specific role?
\item[\textbf{Q3.}] Does MRRG expand the coverage of evaluative
dimensions beyond single-voiced generation, or are its gains simply due
to producing more criteria?
\end{itemize}

\begin{table*}[t!]
\centering
\caption{
Accuracy (\%) performance comparison of different models on RewardBench-2, JudgeBench, and PPE.
S0 denotes SVRG w/o SR, S1 denotes SVRG w/ SR, and CtT denotes Chasing the Tail.
For each model--benchmark pair, the best result is marked in \textbf{bold}.
}
\label{tab:judge-main}
\setlength{\tabcolsep}{4.5pt}
\resizebox{0.9\textwidth}{!}{
\begin{tabular}{l cccc| cccc| cccc}
\toprule
\textbf{Model}
& \multicolumn{4}{c|}{\textbf{RewardBench-2}}
& \multicolumn{4}{c|}{\textbf{JudgeBench}}
& \multicolumn{4}{c}{\textbf{PPE}} \\
\cmidrule(lr){2-5} \cmidrule(lr){6-9} \cmidrule(lr){10-13}
& \textbf{S0} & \textbf{S1} & \textbf{CtT} & \textbf{MRRG}
& \textbf{S0} & \textbf{S1} & \textbf{CtT} & \textbf{MRRG}
& \textbf{S0} & \textbf{S1} & \textbf{CtT} & \textbf{MRRG} \\
\midrule
\multicolumn{13}{l}{\textit{Qwen2.5 Series}} \\
\midrule
Qwen2.5-3B-Instruct
& 29.4 & 40.3 & 21.3 & \textbf{51.2}
& 35.1 & 40.9 & 31.1 & \textbf{55.1}
& 31.2 & 36.3 & 31.3 & \textbf{46.2} \\

Qwen2.5-7B-Instruct
& 37.6 & 50.7 & 36.5 & \textbf{59.2}
& 32.3 & 35.1 & 40.5 & \textbf{56.9}
& 34.0 & 42.3 & 37.5 & \textbf{48.5} \\

Qwen2.5-32B-Instruct
& 42.3 & 54.9 & 59.4 & \textbf{65.3}
& 37.1 & 47.4 & 52.0 & \textbf{65.1}
& 44.0 & 48.2 & 49.5 & \textbf{53.9} \\

\midrule
\multicolumn{13}{l}{\textit{GPT Series}} \\
\midrule
GPT-OSS-20B
& 51.2 & 62.4 & 60.3 & \textbf{73.7}
& 54.3 & 67.7 & 63.1& \textbf{74.0}
& 40.1 & 52.2 &  53.2& \textbf{56.3} \\

GPT-OSS-120B
& 52.3 & 64.4 &  61.3& \textbf{74.5}
& 56.5 & 67.8 & 64.0& \textbf{74.8}
& 43.3 & 52.8 &54.4 & \textbf{57.8} \\

% GPT-4o
% & 49.7 & 58.9 & 58.8 & \textbf{63.0}
% & 46.2 & 42.0 & 48.7 & \textbf{51.1}
% & - & - & - & - \\

\bottomrule
\end{tabular}
}
\end{table*}

\subsection{Experimental Setup}
\label{sec:exp:setup}

\paragraph{Rubric generator and judge models.}
For preference validation experiments, we instantiate $G(\cdot)$ with a diverse range of models including \textbf{Qwen2.5-3B-Instruct, Qwen2.5-7B-Instruct, Qwen2.5-32B-Instruct, GPT-oss-20b and GPT-oss-120b} for rubric generation and  per-criterion
judging. And all methods are evaluated on JudgeBench \cite{judgebench2024}, RewardBench-2 \citep{malik2025rewardbench2advancingreward} and PPE \cite{frick2024evaluate}. As for RLVR experiments, we use \textbf{Qwen2.5-3B-Instruct} as base model for GRPO, while \textbf{GPT-4o} is used for rubric generation for different strategies and \textbf{gpt-oss-120b} is used for per-criterion judging. We select $2000$ questions from Wildchat \cite{zhao2024wildchat} as training set, and models are evaluated on two widely used open-ended-task datasets: BiGGen Bench \cite{kim2024biggenbenchprincipledbenchmark} and HealthBench-Hard \cite{arora2025healthbench}. 

Detailed description for these two datasets is provided in Appendix~\ref{app-data-des}. Also, full prompt templates for all five roles, and the judge template are provided in Appendix~\ref{app-template}.

\paragraph{Baselines.}
We compare MRRG against three methods: 
%\emph{(i) Scalar RM (Skywork-Reward-V2) \cite{liu2025skywork}}: the method generates score directly given the question and response.
\emph{(i) Single-voice rubric generator without sample responses (SVRG w/o SR)}: the method generates rubrics for questions without sample responses. 
\emph{(ii) Single-voice rubric generators with sample responses (SVRG w/ SR)}: the method generates rubrics for specific questions with sample responses.
\emph{(iii) Chasing the Tail\cite{zhang2025chasing}}: the method generates and refines rubrics by comparing great and diverse responses.  

\begin{figure}[h]
    \centering
    \includegraphics[width=1\linewidth]{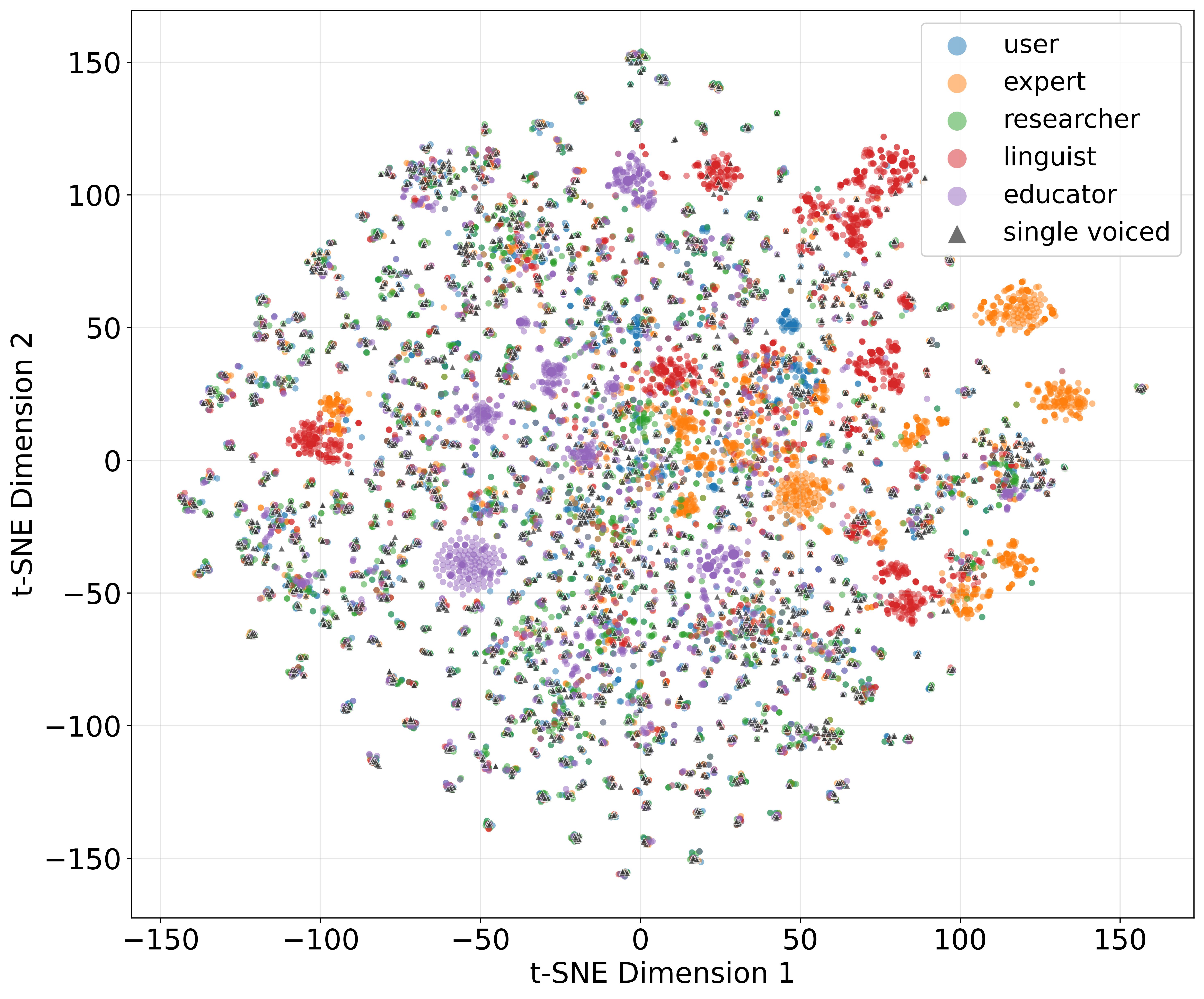}
    \caption{T-SNE visualization of rubrics proposed by different roles and methods.}
    \label{fig:tsne}
\end{figure}

\subsection{Preference Validation}
\label{sec:exp:judge}

Table~\ref{tab:judge-main} shows that MRRG consistently achieves the
strongest performance across all three benchmarks and all backbone models,
often with substantial margins over the baselines. On the Qwen2.5 series,
MRRG improves upon the SVRG w/o SR baseline by more than $20$ points on
RewardBench-2 for the 3B, 7B, and 32B variants, and yields comparable
improvements on JudgeBench. On the GPT series, MRRG remains uniformly
superior, raising GPT-OSS-120B from $52.3$ to $74.5$ on RewardBench-2 and from
$56.5$ to $74.8$ on JudgeBench. These consistent gains across model scales and
families demonstrate that the advantage of multi-role rubric generation is
not tied to a particular backbone. They also suggest that MRRG constructs
more comprehensive and discriminative rubrics for open-ended evaluation.

We also visualize the rubrics generated by different roles and by the
single-voiced process in Figure~\ref{fig:tsne}. The visualization shows
that role-specific rubrics largely cover the regions occupied by
single-voiced rubrics. Meanwhile, they also extend to additional regions
where few or no single-voiced rubrics are present, suggesting that the
multi-role process captures evaluative dimensions that are underexplored
by single-voiced generation. This provides qualitative evidence that
MRRG yields more diverse and comprehensive
criteria.

\begin{figure}[h]
\centering
\subfloat[Single role on JudgeBench]{
    \includegraphics[scale=0.2]{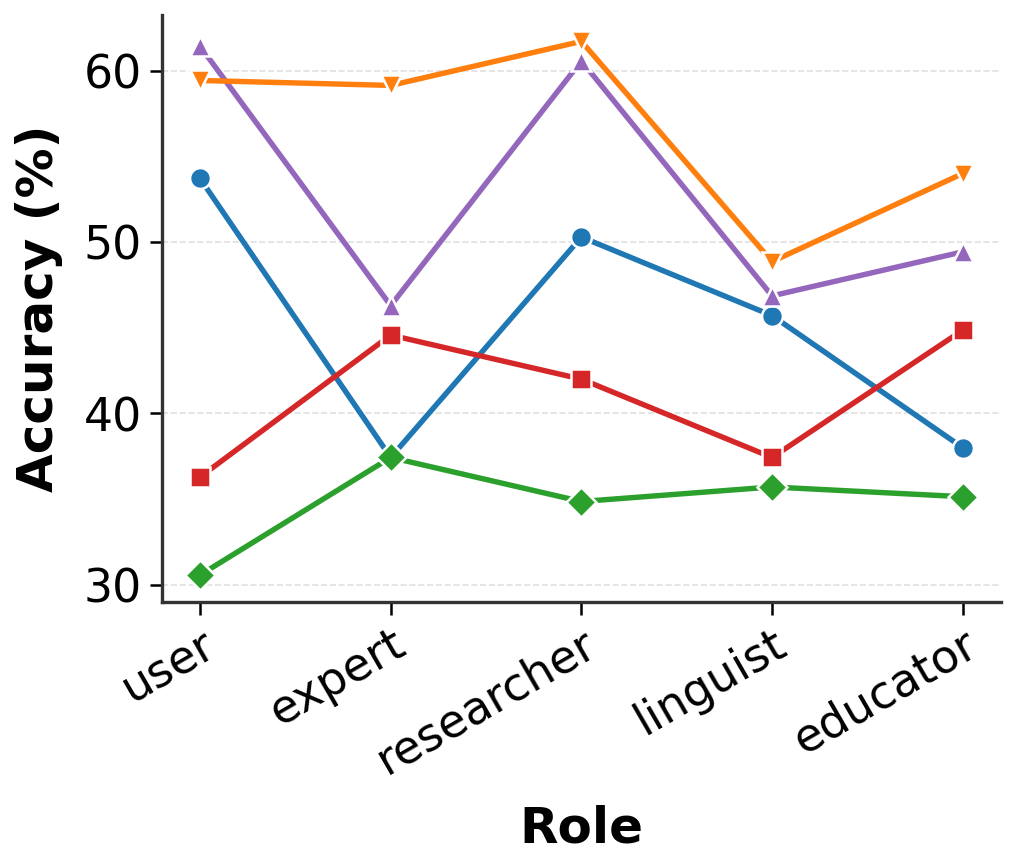}
}
\hfill
\subfloat[leave-one-role-out MRRG on JudgeBench]{
    \includegraphics[scale=0.2]{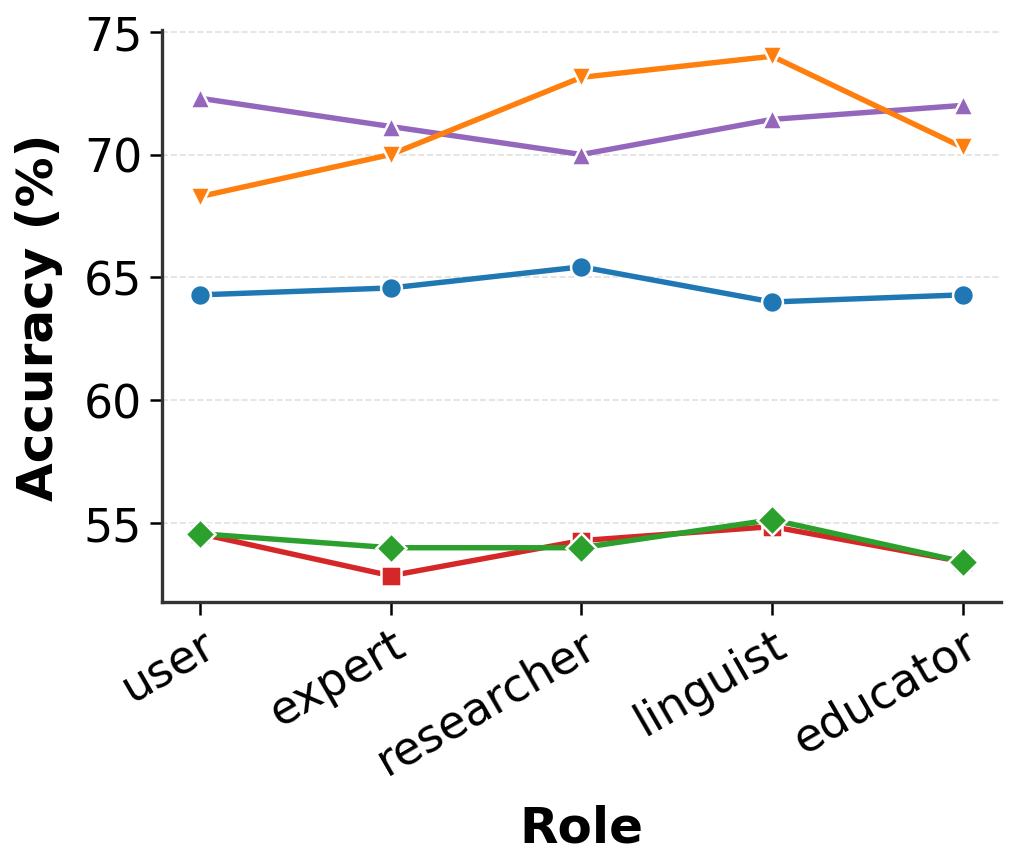}
}
\\
\subfloat[Single role on RewardBench-2]{
    \includegraphics[scale=0.2]{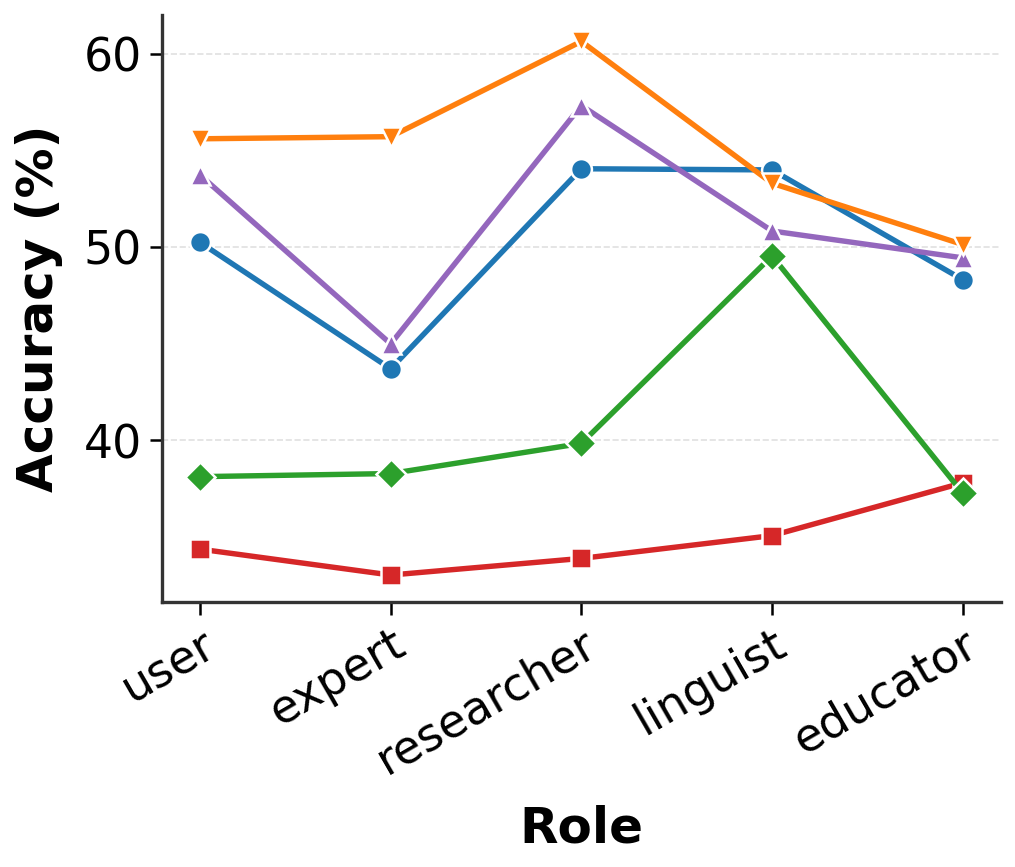}
}
\hfill
\subfloat[leave-one-role-out MRRG on RewardBench-2]{
    \includegraphics[scale=0.2]{judgebench__leave_one_out_no_merge.png}
}\\
\subfloat{
    \includegraphics[scale=0.4]{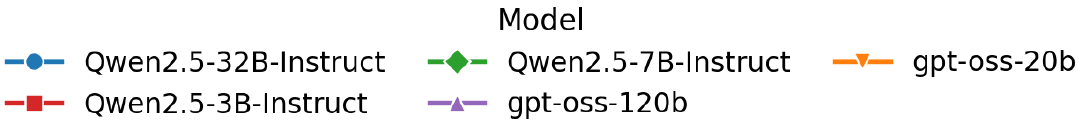}
}
\caption{comparison of single-role
baselines with leave-one-role-out variants of MRRG}
\label{fig:single and leave one}
\vspace{-15pt}
\end{figure}

\paragraph{Ablations.}
\label{sec:ablation}
Table~\ref{tab:judge-ablation} isolates the contribution of different
components of MRRG. \emph{(i) Role ablation}: using rubrics from any
single role leads to a clear performance drop compared with MRRG,
indicating that no individual perspective is sufficient to construct a
comprehensive evaluation rubric. Among the single-role variants,
\textsc{Educator} performs best on both RewardBench-2 and JudgeBench,
achieving 37.8 and 44.8, respectively. Nevertheless, it still lags behind
MRRG by 13.4 points on RewardBench-2 and 10.3 points on JudgeBench,
highlighting the importance of aggregating complementary evaluative
perspectives. \emph{(ii) Repeated single-voiced rubric generation}: To examine whether
the advantage of MRRG simply comes from generating more criteria, we
introduce a repeated single-voiced rubric generation baseline, denoted as
RSVRG. Specifically, RSVRG repeats the single-voiced rubric generation
process $5$ times and then applies the same post-processing pipeline as
MRRG. As shown in Table~\ref{tab:judge-ablation}, RSVRG improves over
individual-role variants in some cases, especially when sample responses
are enabled, suggesting that repeated sampling can increase rubric
diversity to a limited extent. However, RSVRG w/ SR remains substantially
below MRRG, with gaps of $7.3$ points on RewardBench-2 and $10.0$ points
on JudgeBench. This indicates that simply generating more single-voiced
rubrics cannot fully substitute for explicitly eliciting rubrics from
multiple complementary roles. Overall, these results suggest that the
gains of MRRG arise from role-based diversity rather than merely from an
increased number of generated criteria.

In addition, we conducted an experiment to compare single-role
baselines with leave-one-role-out MRRG variants, as shown in
Figure~\ref{fig:single and leave one}. The results show that the most
effective single role varies across backbone models, suggesting that no
single evaluative perspective is uniformly optimal. In contrast, removing
a role from MRRG leads only to moderate performance changes in most cases. Although
excluding stronger single roles tends to result in larger drops, the
overall degradation remains limited. This indicates that MRRG does not
depend critically on any particular role; rather, its effectiveness comes
from aggregating complementary perspectives to construct more
comprehensive rubrics.

\begin{table}[h]
\centering
\small
\caption{Ablation of MRRG components using \textbf{Qwen2.5-3B-Instruct} on the domain benchmark. ``role $m$'' denotes using only rubrics from role $m$.}
\label{tab:judge-ablation}
\resizebox{0.45\textwidth}{!}{
\begin{tabular}{lcc}
\toprule
Variant & RewardBench-2 & JudgeBench \\
\midrule
\textsc{User}      &34.4&36.2                  \\
 \textsc{Domain Expert}  &33.0&44.5             \\
 \textsc{Educator}        &37.8& 44.8             \\
 \textsc{Researcher}       &33.9&  42.0          \\
 \textsc{Linguist}          &35.1& 37.4       \\
\midrule
RSVRG w/o SR  &39.1&38.5\\
RSVRG w/ SR  &43.9&45.1\\
\midrule
MRRG  &51.2 &55.1   \\

\bottomrule
\end{tabular}
}
\end{table}

% \begin{figure}[t]
% \centering
% \includegraphics[width=\linewidth]{figs/kappa_rho_delta.pdf}
% \caption{Empirical estimates of $\kappa$, $\rho$, and $\delta$
% across benchmarks, and the ratio of realized MSE to the
% theoretical bound of Eq.~\eqref{eq:end-to-end}.}
% \label{fig:kappa-rho}
% \end{figure}

\subsection{Reward Modeling for RLVR}
\label{sec:exp:rlvr}

\paragraph{Setup.}
We plug $S(x,y;\widetilde{R}(x))$ into GRPO as the per-rollout
reward, using the same rubric $\widetilde{R}(x)$ cached per prompt
across the rollout group. The base policy is
\textbf{Qwen2.5-3B-Instruct}; we train it on $2000$ open-ended prompts drawn from \textbf{WildChat} \cite{zhao2024wildchat}.
%and more details related to hyperparameters are listed in Appendix~\ref{app:impl}.

\paragraph{Main results.}
Table~\ref{tab:rlvr-main} reports the post-RL policy performance under
different reward sources. Using MRRG as the reward consistently yields the
best results on both in-domain and domain-transfer evaluations. On BiGGen
Bench, MRRG achieves a score of $63.7$, outperforming the strongest
baseline, SVRG w/ SR, by $1.7$ points. The improvement is even larger on
the domain-transfer benchmark HealthBench-Hard, where MRRG reaches $32.1$
and exceeds the second-best method by $3.4$ points. Notably, all
rubric-based rewards improve over the base policy, while MRRG provides the
largest gains across both benchmarks. These results suggest that
multi-role rubric rewards provide a more comprehensive and robust training
signal than single-voiced alternatives, leading to better generalization
under distribution shift. Detailed experimental results are provided in
Appendix~\ref{app-detail beggen}.

\begin{table}[t]
\centering
\small
\caption{Post-RL policy Accuracy (\%) performance under different reward sources.
 Best performance in each column is 
in \textbf{bold}.}
\label{tab:rlvr-main}
\resizebox{0.47\textwidth}{!}{
\begin{tabular}{lcc}
\toprule
Reward source  & BiGGen Bench & HealthBench-Hard  \\
\midrule
Base policy (Qwen2.5-3B-Instruct)               & 57.8 & 25.6 \\
SVRG w/o SR                       & 60.0 & 26.0 \\
SVRG w/ SR                   & 62.0 & 28.7  \\
Chasing the Tail         & 61.1 & 27.0 \\
\midrule
MRRG               & \textbf{63.7} & \textbf{32.1}  \\
\bottomrule
\end{tabular}
}
\end{table}

% \begin{figure}[t]
% \centering
% \includegraphics[width=0.7\linewidth]{figs/hacking.pdf}
% \caption{Reward inflation under single-axis adversarial responses.
% Dashed line marks the $|\mathcal{S}|/M = 1/5$ theoretical bound of
% Proposition~\ref{prop:hacking} for \methodname{} with $M=5$.}
% \label{fig:hacking}
% \end{figure}

%\subsection{Qualitative Analysis}
%\label{sec:exp:case}

%Figure~\ref{fig:case-study} shows two representative prompts with
%their per-persona rubrics and the consolidated
%$\widetilde{R}(x)$. In the \textbf{[medical-diagnosis]} example,
%\textsc{Domain Expert} surfaces a safety-critical criterion that no
%other persona identifies, and consolidation retains it with weight
%$3$; in the \textbf{[creative-writing]} example,
%\textsc{Linguist} and \textsc{Educator} each contribute stylistic
%criteria that would be absent under a single-voice generator. These
%cases illustrate how the five axes produce genuinely complementary
%coverage rather than rephrased duplicates.

\section{Related Work}
\label{sec:related}

\subsection{Reward Modeling and LLM-as-a-Judge}
\textbf{Scalar reward models.} Aligning LLMs with human intent typically relies on a learned reward model
trained on Bradley--Terry-style pairwise preferences. InstructGPT~\cite{ouyang2022training} and HH-RLHF \cite{bai2022training} train a Bradley--Terry head on top of a base language model (LM) using
human-labeled pairwise preferences, and use the resulting scalar as the
reward for RLHF. PPE~\cite{frick2024evaluate} subsequently shows that such reward models are brittle proxies of downstream policy quality and proposes a
preference-proxy benchmark to measure that gap.
\textbf{LLM-as-a-judge.} Other methods like MT-Bench / Chatbot Arena~\cite{zheng2023judging} prompts a strong
LLM (GPT-4) to directly score or pairwise-compare two responses, and
AlpacaFarm~\cite{dubois2023alpacafarm} uses this LLM-judge signal as a stand-in for human feedback when simulating RLHF.  
%Our work targets exactly this entanglement by replacing the implicit signal with an explicit, auditable rubric.

\subsection{Rubric-Based Evaluation and  RLVR}

A recent line of work makes evaluation criteria explicit through rubrics
or checklists. These methods typically generate rubrics using one or a few
LLM calls under a single ``universal evaluator'' prompt, and the resulting
criteria can further be used as verifiable rewards for RLVR. For example,
Chasing the Tail~\citep{zhang2025chasing} emphasizes that rubric
generation should account for distinctions among high-quality and diverse
responses, and proposes a workflow for constructing rubrics that capture
such distinctions. Rubrics as Rewards (RaR)~\citep{gunjal2025rubrics}
generates question-specific rubrics from the input question and a
reference answer, and uses them as rewards for RLVR. OpenRubrics
~\citep{liu2025openrubrics} collects large-scale and diverse
question--rubric pairs to train rubric-generation models capable of
producing reliable and scalable rubrics.

While these approaches demonstrate the effectiveness of rubric-based
evaluation and reward modeling, they often depend on additional
supervision, such as reference answers for rubric construction or
preference-labeled data for training rubric generators. This reliance can
limit their applicability in realistic settings where reference answers
or preference annotations are unavailable or expensive to obtain.

\subsection{Multi-Role Prompting for Evaluation and Reasoning}

Prior work has shown that role conditioning can substantially affect the
knowledge and reasoning patterns elicited from LLMs. In-Context Impersonation \cite{salewski2023context} shows that asking a  LLM to act as a domain
expert / a child  systematically changes the knowledge
it surfaces on QA and bandit tasks; Role-Play Prompting extends role-play to
chain-of-thought reasoning and reports gains across MMLU and GSM8K.
Multi-Agent Debate \cite{chan2024chateval} runs several LLM instances that
iteratively critique and revise each other's answers, improving
factuality on TriviaQA-style tasks; Encouraging Divergent Thinking (EDT) \cite{liang2024encouraging} shows that explicitly
encouraging divergent viewpoints across agents further reduces
mode collapse in reasoning.  These methods primarily use multiple roles to produce better
answers. In contrast, MRRG generates the evaluation
criteria themselves. Moreover, unlike debate-based methods, MRRG does not
require iterative interaction, critique, or consensus among agents.
Instead, role-specific rubrics are generated independently and then
combined through deterministic deduplication.

\section{Conclusion}
\label{sec:conclusion}

We presented \textbf{Multi-Role Rubric Generation (MRRG)}, a
training-free framework that
recasts rubric construction as a multi-role elicitation process and
pool per-role rubrics into a powerful scorer. The
same scorer serves uniformly as an LLM-as-a-judge for preference
validation and as a reward inside RLVR, with no architectural change
between the two uses.  Empirically, MRRG consistently outperforms  single-voice rubric generators, and
rubric-refinement baselines on JudgeBench, RewardBench-2, and PPE,
and it also delivers stronger downstream policies under GRPO on BiGGen-Bench and
HealthBench-Hard. 

\section*{Limitations}

While MRRG mitigates the dimensional blind spots of single-voiced rubric generation, it still has one notable limitation. The pool of roles used to instantiate the multiple voices is predefined and remains fixed across all questions. In practice, however, the set of perspectives that are truly relevant for judging a response is inherently question-dependent: a coding task and a moral-reasoning task, for instance, call for largely disjoint expert viewpoints. Using a static role pool may therefore introduce uninformative voices on some questions while omitting critical ones on others. A natural extension is to incorporate adaptive, question-conditioned role selection — for example, by retrieving or generating roles on the fly based on the input — which is an
important direction for future work.


\bibliography{custom}
\newpage
\appendix

\section{Dataset description}
\label{app-data-des}
\begin{itemize}
    \item \textbf{JudgeBench \cite{judgebench2024}:} Using a novel pipeline that transforms any dataset with ground truth labels and verification algorithms into a corresponding dataset specifically tailored for LLM-based judges, It constructs a challenging dataset consisting of 350 response pairs across four categories:general knowledge, reasoning, mathematics, and coding. Each pair contains one objectively correct
response and one objectively incorrect response, with the incorrect response designed to contain
subtle errors, making it difficult for LLM-based judges to distinguish between the two.
\item \textbf{RewardBench-2 \cite{malik2025rewardbench2advancingreward}:} The benchmark contains 1865 preference validation data containing the following categories: factuality, precise instruction following, math, safety, focus and ties. Each instance contains at least one correct answer and three wrong answers. For each sample, the judgment is identified as correct only if all the chosen answers obtain higher score than all the rejected answers. 
\item \textbf{PPE: \cite{frick2024evaluate}} Preference Proxy Evaluations is an evaluation benchmark for reward models and LLM-judges, which are used to replicate human preferences for large-scale LLM training and evaluation. PPE contains more than 16000 real human preference data sourced from Chatbot Arena as well as verifiable correctness preference data sourced from reputable LLM benchmarks, such as MMLU-Pro and MATH. PPE provides various metrics across the benchmarks, and includes analysis on which metrics correlated the most with post-RLHF outcomes. PPE is correlated to downstream outcomes to ensure using PPE as a reference evaluation signal when tuning reward models and LLM Judges leads to tangible downstream improvements in post-RLHF LLM performance.
\item \textbf{WildChat: \cite{zhao2024wildchat}} WildChat is a collection of 650K conversations between human users and ChatGPT. The authors collected WildChat by offering online users free access to OpenAI's GPT-3.5 and GPT-4. The dataset contains a broad spectrum of user-chatbot interactions that are not previously covered by other instruction fine-tuning datasets: for example, interactions include ambiguous user requests, code-switching, topic-switching, political discussions, etc. WildChat can serve both as a dataset for instructional fine-tuning and as a valuable resource for studying user behaviors.
\item \textbf{BiGGen Bench: \cite{kim2024biggenbenchprincipledbenchmark}} BIGGEN BENCH evaluates 9 core capabilities of LMs–namely instruction following, grounding, planning, reasoning, refinement, safety, theory of mind, tool usage, and multilingualism–across 77 tasks and 765 instances.
\item \textbf{HealthBench-Hard \cite{arora2025healthbench}:} A subset of 1,000 HealthBench examples chosen because they are difficult for current frontier models. HealthBench Hard is designed to be a difficult target for future models, with many current models achieving a score of zero on it.
\end{itemize}

\section{Implementation Details}
All experiments were conducted on a system equipped with approximately 500 GB of GPU memory, with GRPO training taking approximately 7 hours. Detailed hyperparameters for GRPO are provided in Table \ref{tab:grpo_hyperparams}.

\begin{table}[htbp]
\centering
\small
\caption{Hyperparameters for GRPO training.}
\label{tab:grpo_hyperparams}
\resizebox{0.48\textwidth}{!}{
\begin{tabular}{ll}
\hline
\textbf{Hyperparameter} & \textbf{Value} \\
\hline
Base model & \texttt{Qwen/Qwen2.5-3B-Instruct} \\
Use PEFT & \texttt{False} \\
\hline
Number of rollout & $8$ \\
Temperature & $1.0$ \\
Top-$p$ & $1.0$ \\
Top-$k$ & $-1$ \\
$\beta$ &$0.001$\\
\hline
Learning rate & $1 \times 10^{-6}$ \\
Batch size per device & $8$ \\
Gradient accumulation steps & $4$ \\
Number of epochs & $3$ \\
Warmup ratio & $0.1$ \\
LR scheduler & \texttt{constant with warmup} \\
Optimizer & \texttt{adamw} \\
\hline
\end{tabular}
}
\end{table}

\section{Detailed experimental results}
\label{app-experiment}
\subsection{Experimental judging  results in different domains}
\label{app-experiment-domain}
\begin{figure*}[t!]
\centering
\subfloat[Qwen2.5-3b-Instruct]{%
    \includegraphics[scale=0.21]{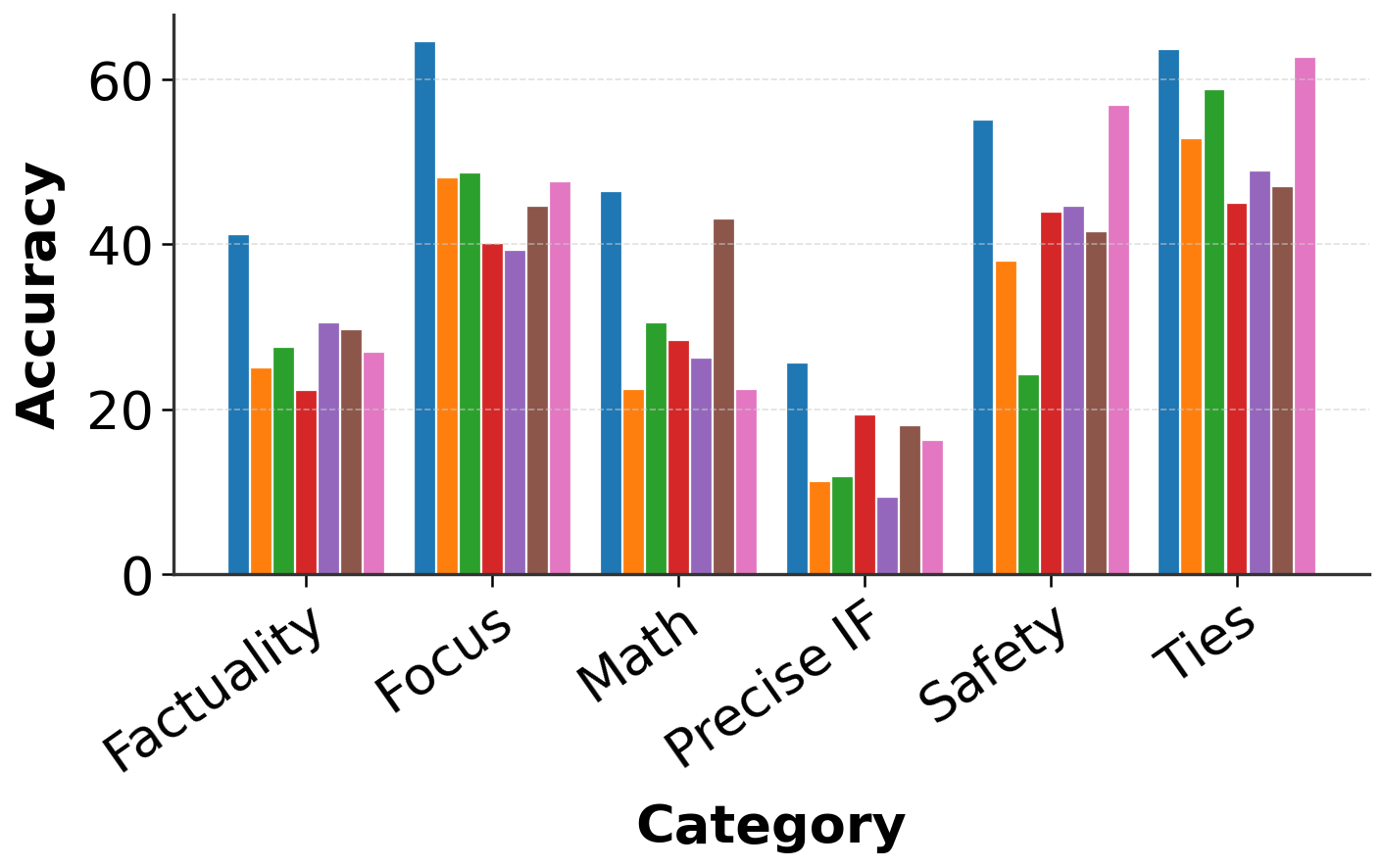}
}
\hfill
\subfloat[Qwen2.5-7b-Instruct]{%
    \includegraphics[scale=0.21]{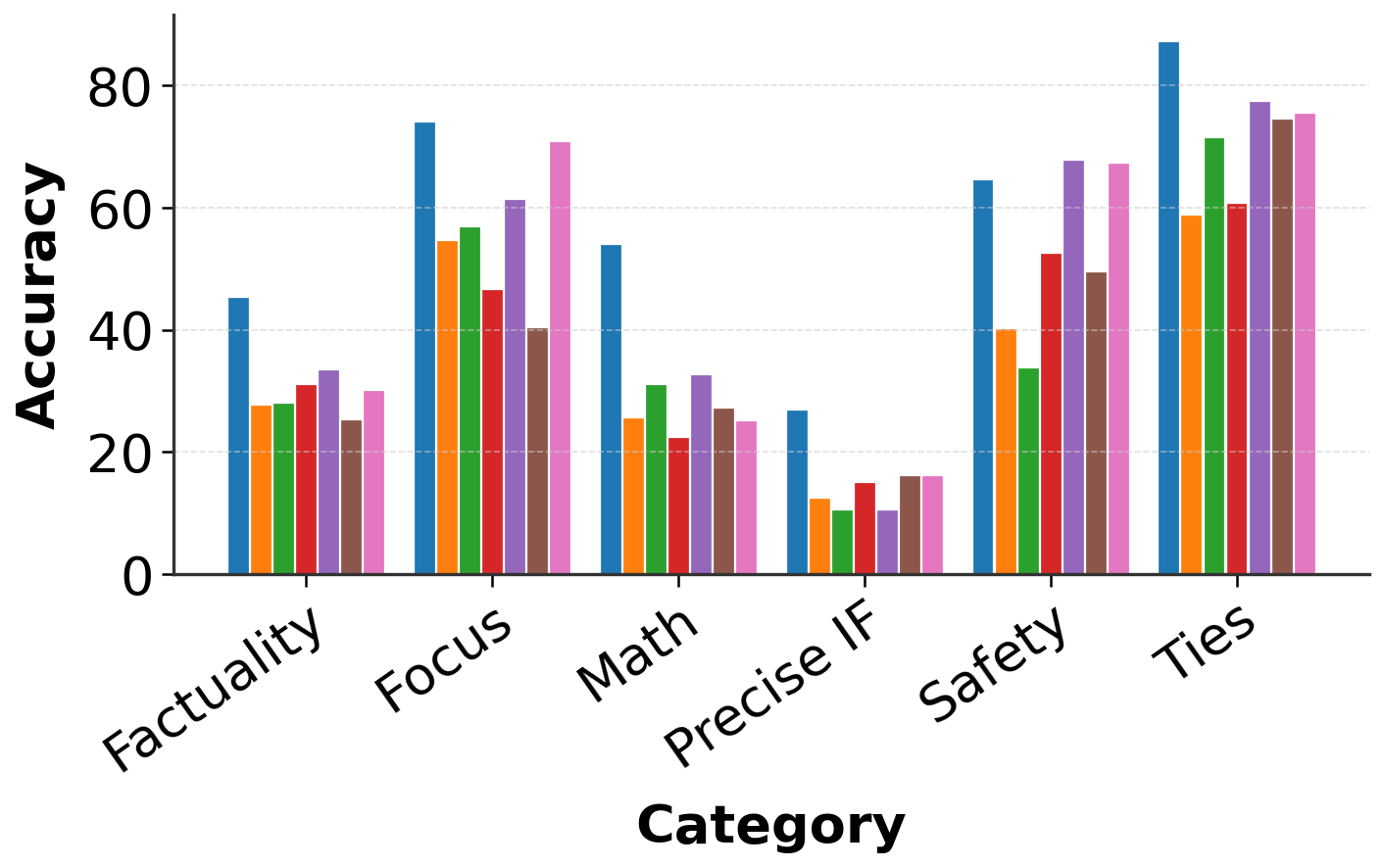}
}
\hfill
\subfloat[Qwen2.5-32b-Instruct]{%
    \includegraphics[scale=0.21]{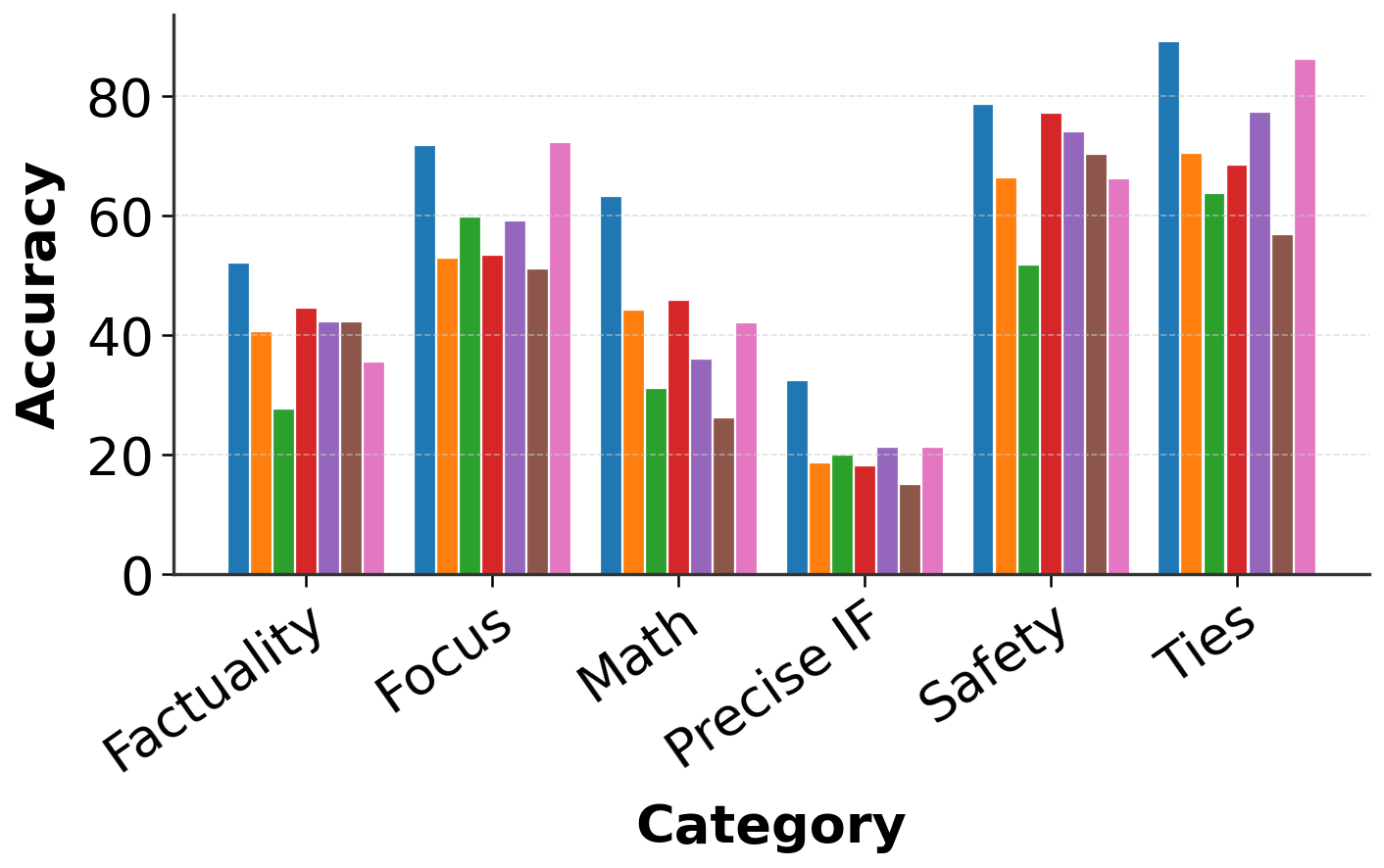}
}

\par\vspace{4pt}

\makebox[\linewidth][c]{%
    \subfloat[GPT-OSS20b]{%
        \includegraphics[scale=0.21]{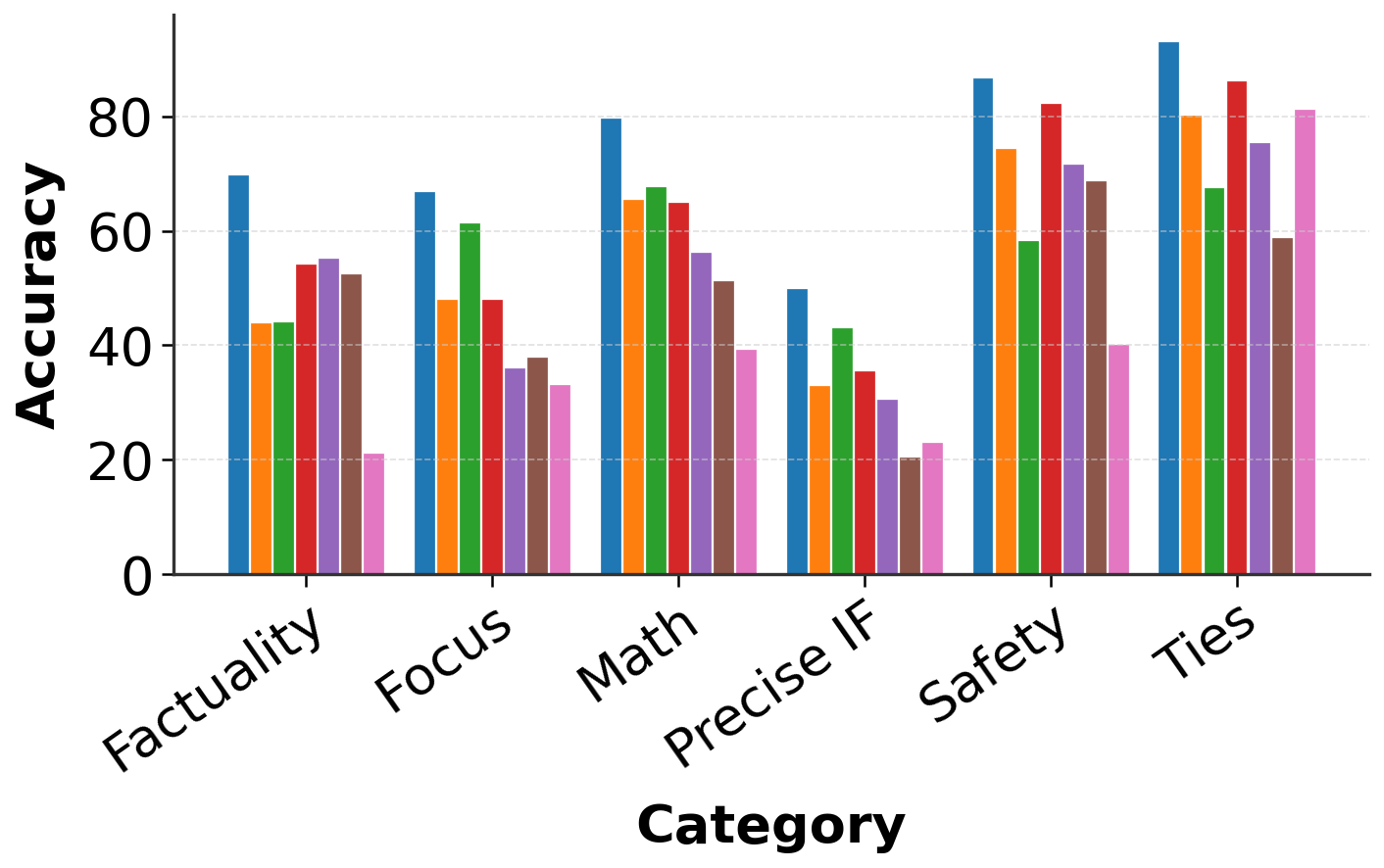}
    }
    \hspace{0.08\linewidth}
    \subfloat[GPT-OSS120b]{%
        \includegraphics[scale=0.21]{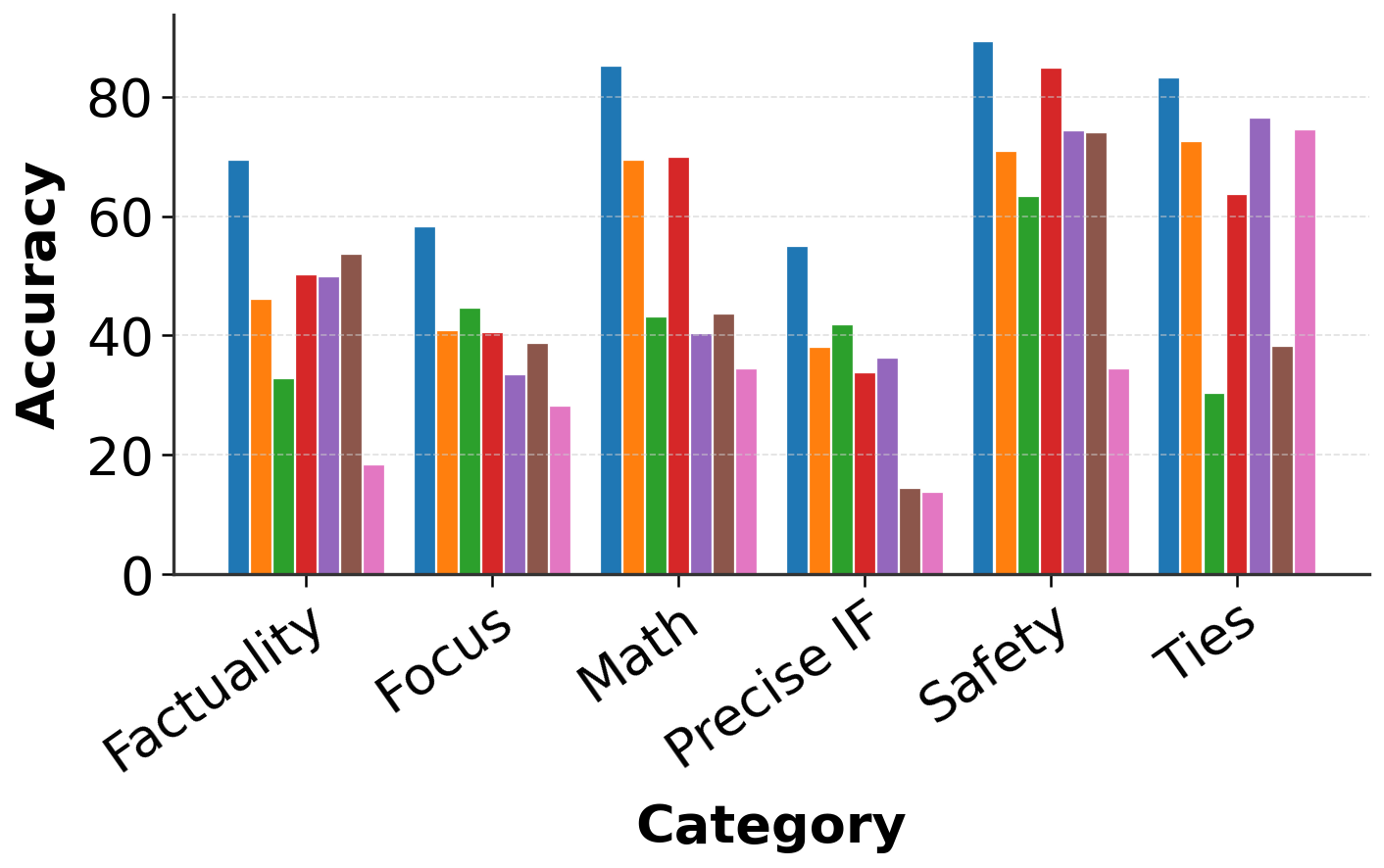}
    }
}

\par\vspace{2pt}

\subfloat{%
    \includegraphics[scale=0.7]{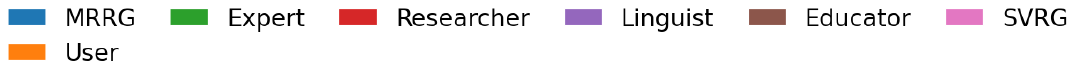}
}
\caption{Comparison of different methods on each domain from RewardBench-2.}
\label{fig:category reward}
\end{figure*}

\begin{figure*}[t]
\centering
\subfloat[GPT-OSS-20b]{
    \includegraphics[scale=0.25]{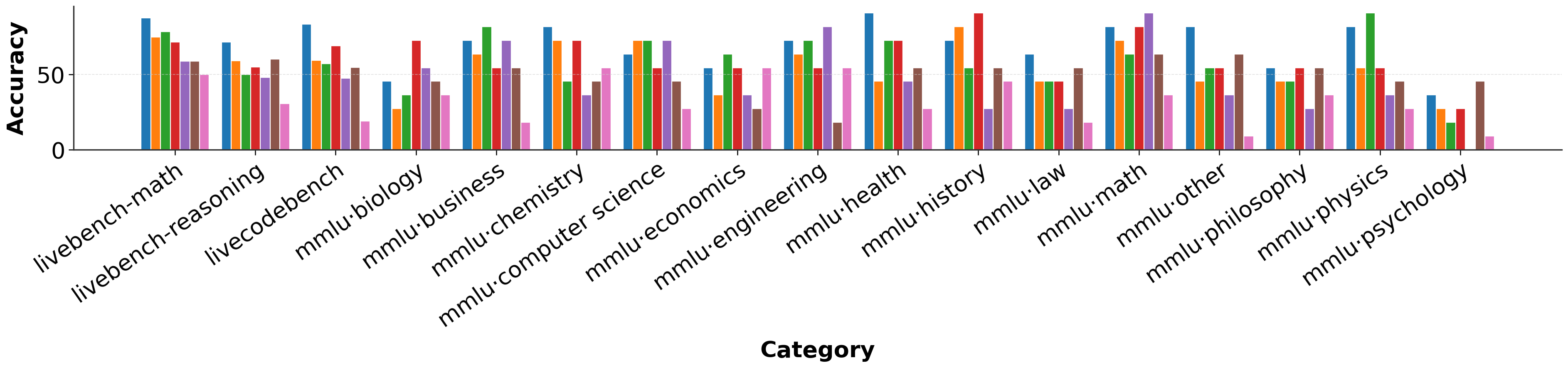}
}
\\
\subfloat[GPT-OSS-120b]{
    \includegraphics[scale=0.25]{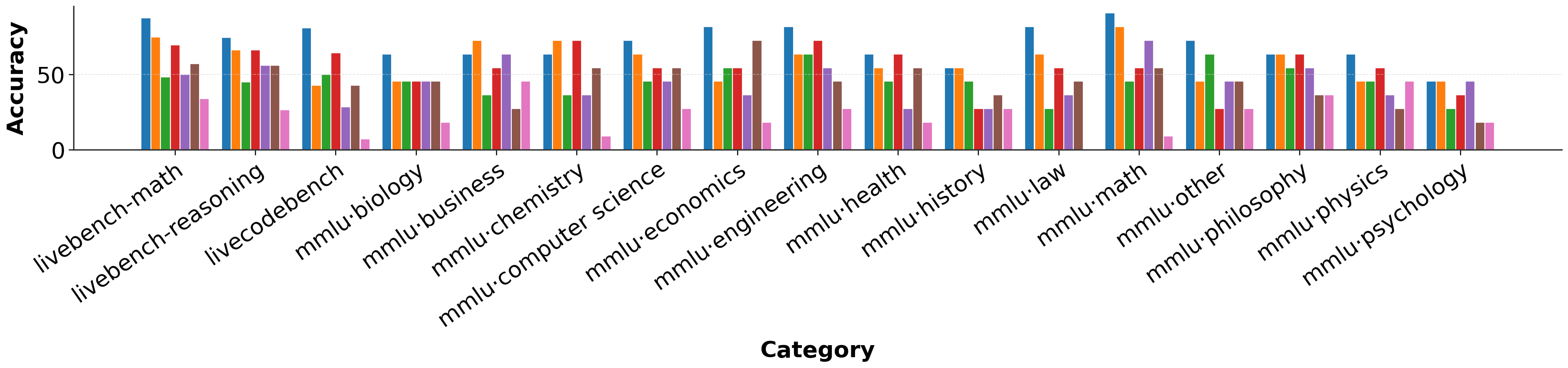}
}
\\
\subfloat[Qwen2.5-3b-Instruct]{
    \includegraphics[scale=0.25]{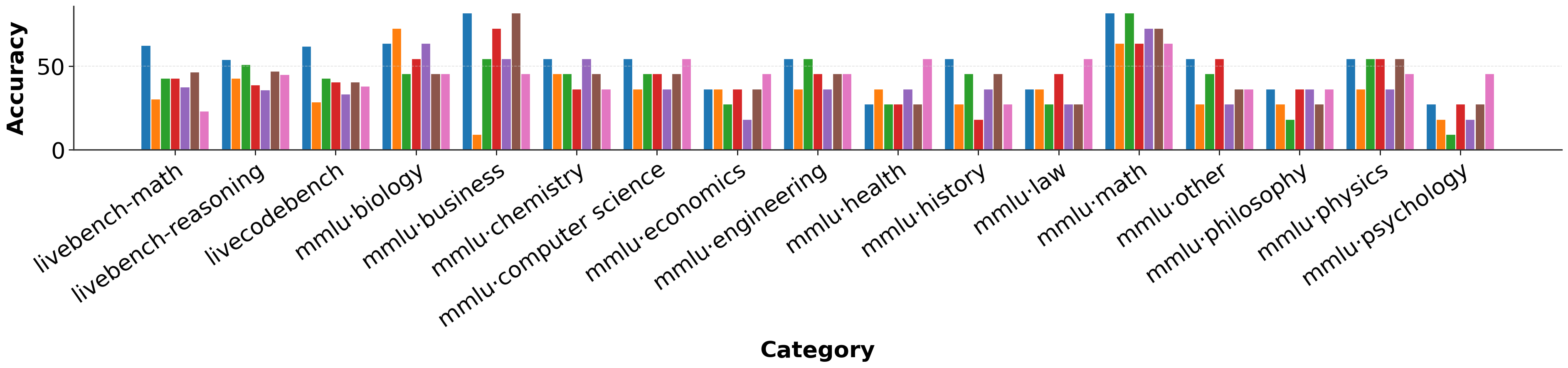}
}
\\
\subfloat[Qwen2.5-32b-Instruct]{
    \includegraphics[scale=0.25]{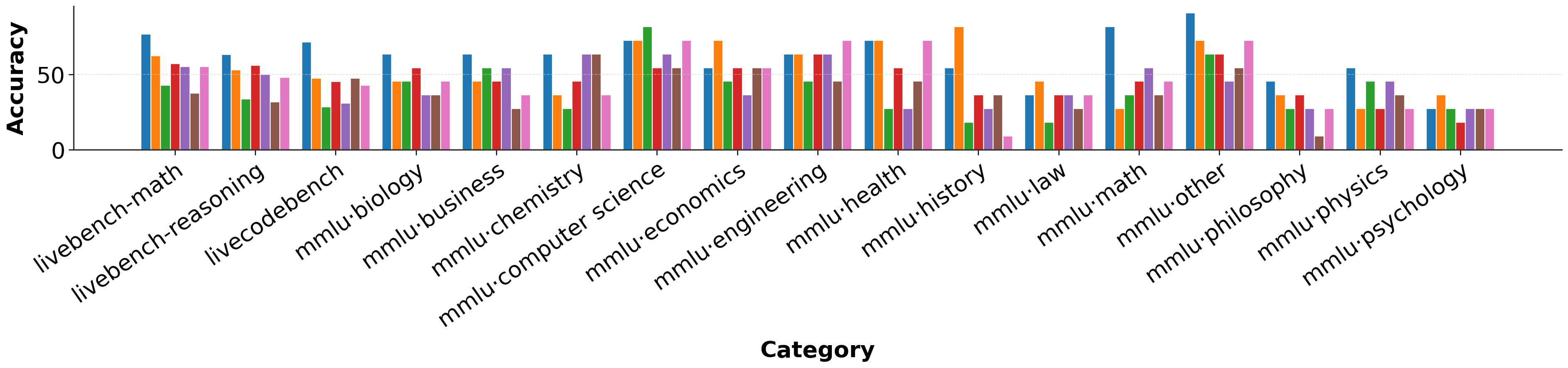}
}
\\
\subfloat{
    \includegraphics[scale=0.8]{reward_legend.pdf}
}
\caption{comparison of different methods on each domain from JudgeBench.}
\label{fig:category judge}
%\vspace{-15pt}
\end{figure*}

In Figure \ref{fig:sub_acc}, we report the experimental results of Qwen2.5-7b-Instruct on each domain from JudgeBench. In this subsection, we report the performance of each method on each domain from RewardBench-2 and JudgeBench in Figure \ref{fig:category reward} and Figure \ref{fig:category judge}. From the figures, we can observe domain-dependent judging accuracy in single-voiced rubric and other single-role rubric caused dimensional blind spots.

% \begin{figure*}[t!]
% \centering
% \subfloat[Qwen2.5-3b-Instruct]{
%     \includegraphics[scale=0.21]{reward_bench__Qwen2.5-3B-Instruct.png}
% }
% \hfill
% \subfloat[Qwen2.5-7b-Instruct]{
%     \includegraphics[scale=0.21]{reward_bench__Qwen2.5-7B-Instruct.png}
% }
% \hfill
% \subfloat[Qwen2.5-32b-Instruct]{
%     \includegraphics[scale=0.21]{reward_bench__Qwen2.5-32B-Instruct.png}
% }
% \\
% \subfloat[GPT-OSS20b]{
%     \includegraphics[scale=0.21]{reward_bench__gpt-oss-20b.png}
% }
% \hfill
% \subfloat[GPT-OSS120b]{
%     \includegraphics[scale=0.21]{reward_bench__gpt-oss-120b.png}
% }
% \\
% \subfloat{
%     \includegraphics[scale=0.7]{reward_legend.pdf}
% }
% \caption{comparison of different methods on each domain from RewardBench-2.}
% \label{fig:category reward}
% %\vspace{-15pt}
% \end{figure*}

\subsection{Experimental Setup for Rubric-Hacking Experiments}
\label{sec:rubric_hacking_setup}

Figure~\ref{fig:sub_rubric} reports our rubric-hacking experiments. The
experimental procedure is as follows. We first sample $400$ query prompts
and use Qwen2.5-3B-Instruct to generate both single-voiced and
multi-voiced rubrics for each query. Given each generated rubric, we then
prompt GPT-4o to produce an adversarial answer that attempts to satisfy
the rubric while remaining low-quality along evaluative dimensions not
covered by the rubric. The prompt template used for this adversarial
answer generation is shown in
Figure~\ref{fig:rubric-passing answer prompt_template}. Finally, GPT-4o
is asked to evaluate the generated answers on a $0$--$10$ scale using the
prompt template in Figure~\ref{fig:answer evaluation prompt}.

The motivation behind this experiment is that a more comprehensive rubric
should be harder to hack. If a rubric covers a broader set of evaluative
dimensions, it becomes more difficult for an adversary to generate an
answer that both satisfies the rubric and remains low-quality by
exploiting dimensional blind spots. Conversely, if such a low-quality yet
rubric-passing answer can be easily generated, this suggests that the
rubric fails to capture important aspects of answer quality. Therefore,
the quality of these adversarial rubric-passing answers serves as an
indirect measure of rubric comprehensiveness and robustness: higher
evaluated answer quality indicates that the rubric leaves fewer exploitable
blind spots, whereas lower quality indicates greater vulnerability to
rubric hacking.

\subsection{Ablation on LLM-Based Rubric Consolidation}
\label{sec:ablation_consolidation}

In this subsection, we empirically examine whether an additional
LLM-based consolidation step benefits MRRG. Specifically, after generating
rubrics with the multi-role process, we prompt an LLM to consolidate them
by merging semantically overlapping criteria and removing mutually
conflicting ones using prompt in Figure \ref{fig:CONSOLIDATION prompt_template}. This ablation allows us to assess whether semantic
rubric refinement improves upon our default exact-match deduplication
strategy.

\begin{table}[h!]
\centering
\caption{
Ablation on LLM-based rubric consolidation. MRRGC denotes MRRG with LLM-based
consolidation. For each model--benchmark pair, the better result is marked in
\textbf{bold}.
}
\label{tab:judge_ab_consolidation}
\setlength{\tabcolsep}{6pt}
\resizebox{0.45\textwidth}{!}{
\begin{tabular}{l cc cc}
\toprule
\textbf{Model}
& \multicolumn{2}{c}{\textbf{RewardBench-2}}
& \multicolumn{2}{c}{\textbf{JudgeBench}} \\
\cmidrule(lr){2-3}
\cmidrule(lr){4-5}
& \textbf{MRRGC} & \textbf{MRRG}
& \textbf{MRRGC} & \textbf{MRRG} \\
\midrule
\multicolumn{5}{l}{\textit{Qwen2.5 Series}} \\
\midrule
Qwen2.5-3B-Instruct
& 46.2 & \textbf{51.2}
& 47.1 & \textbf{55.1} \\

Qwen2.5-7B-Instruct
& 52.7 & \textbf{59.2}
& 43.7 & \textbf{56.9} \\

Qwen2.5-32B-Instruct
& 59.2 & \textbf{65.3}
& 57.1 & \textbf{65.1} \\

\bottomrule
\end{tabular}
}
\end{table}

\begin{table*}[t]
\centering
\small
\caption{Detailed post-RL policy performance on BiGGen Bench under different reward sources. Best annotation-free reward per column in \textbf{bold}.}
\label{tab:rlvr-biggen}
\resizebox{\textwidth}{!}{%
\begin{tabular}{lcccccccccc}
\toprule
Reward source  & grounding & instruction following & multilingual & planning & reasoning & refinement & safety & theory of mind & tool usage & average \\
\midrule
Base policy (Qwen2.5-3B-Instruct) & 72.5 & 67.5 & 32.9 & 52.1 & 68.2 & 54.6 & 61.7 & 62.2 & 30.7 & 57.8 \\
%Scalar RM                         & -- & -- & -- & -- & -- & -- & -- & -- & -- & -- \\
SVRG w/o SR        & 73.0 & 69.8 & 35.0 & 58.6 & 70.0 & 57.9 & 56.7 & 64.8 & 38.6 & 60.0 \\
SVRG w/ SR         & 73.3 & \textbf{72.0} & 32.1 & \textbf{60.0} & 72.3 & 63.2 & 61.1 & 67.8 & 40.4 & 62.0 \\
Chasing the Tail                  & 73.3 & 71.3 & 34.3 & 59.3 & 72.5 & 60.2 & 55.7 & 66.8 & 40.7 & 61.1 \\
\midrule
MRRG & \textbf{74.0} & 70.2 & \textbf{38.9} & 57.9 & \textbf{74.3} & \textbf{66.1} & \textbf{66.7} & \textbf{68.2} & \textbf{42.1} & \textbf{63.7} \\
\bottomrule
\end{tabular}%
}
\end{table*}

\begin{table*}[t]
\centering
\small
\caption{Detailed post-RL policy performance on HealthBench-Hard under different reward sources. Best annotation-free reward per column in \textbf{bold}.}
\label{tab:rlvr-health}
\resizebox{\textwidth}{!}{%
\begin{tabular}{lcccccc}
\toprule
Reward source  & accuracy & communication quality & completeness & context awareness & instruction following & average \\
\midrule
Base policy (Qwen2.5-3B-Instruct) & 21.5 & 40.6 & 14.9 & 16.9 & 34.0 & 25.6 \\
%Scalar RM                        &      &      &      &      &      &      \\
SVRG w/o SR        & 22.9 & 40.0 & 17.1 & 18.9 & 30.9 & 26.0 \\
SVRG w/ SR         & \textbf{25.8} & 43.0 & \textbf{19.5} & 19.5 & 35.5 & 28.7 \\
Chasing the Tail                  & 21.9 & 48.7 & 15.3 & 13.1 & 35.9 & 27.0 \\
\midrule
MRRG & 24.3 & \textbf{55.1} & 18.7 & \textbf{20.5} & \textbf{42.0} & \textbf{32.1} \\
\bottomrule
\end{tabular}%
}
\end{table*}

As shown in Table~\ref{tab:judge_ab_consolidation}, LLM-based
consolidation does not improve MRRG. Instead, MRRG consistently
outperforms MRRGC across all backbone models and both benchmarks. On
RewardBench-2, MRRG exceeds MRRGC by $5.0$ to $15.0$ points, while on
JudgeBench the gains range from $8.0$ to $14.6$ points. The degradation is
particularly pronounced for the GPT-OSS models, where consolidation leads
to more than $12$-point drops on both benchmarks.

These results suggest that semantic consolidation is not a free
improvement for rubric generation. Although consolidation can reduce
surface-level redundancy, it may also remove useful role-specific criteria
or suppress legitimate differences between evaluative perspectives.
Moreover, the LLM used for consolidation must itself decide which criteria
are redundant or conflicting, introducing an additional source of
subjective bias. We therefore adopt exact-match deduplication as the
default aggregation strategy, as it preserves complementary multi-role
criteria while remaining deterministic and reproducible.

\subsection{Detailed experimental results on BigGen Bench and Health-Bench-Hard}
\label{app-detail beggen}
In this subsection, we provide detailed post-RL policy performance on BiGGen Bench and HealthBench-Hard under different reward sources in Table \ref{tab:rlvr-biggen} and \ref{tab:rlvr-health} respectively.

\subsection{More case studies}

Table~\ref{tab:huli_answer_comparison} illustrates a case where the single-voiced rubric fails to separate a well-grounded answer from a superficially plausible one. The query ``What is huli?'' is inherently ambiguous, since the term may refer to a Hawaiian verb, the food-related expression ``huli huli'', or other language-specific meanings. The rejected answer appears fluent and informative, but it adds weakly supported associations such as mischief, idiomatic usage, and unrelated food references. Because the single-voiced rubric mainly checks whether the response defines the term, provides context, and includes examples, it assigns the rejected answer the same score as the chosen answer. By contrast, the multi-voiced rubric includes more discriminative criteria concerning ambiguity resolution, calibrated uncertainty, precise wording, and avoidance of unsupported or tangential details. These criteria allow MRRG to prefer the chosen answer, which presents the main plausible meanings, explains the relevant usage of ``huli huli'', and appropriately asks for additional context. This example demonstrates that MRRG better captures factual reliability under ambiguity, where the key challenge is not producing more information, but selecting information that is grounded, relevant, and properly scoped.

\section{Prompt template for rubric generation and evaluation}
\label{app-template}
In this section, we provide all the prompting template in our experiments in the following figures:
\begin{itemize}
    \item Rubric passing answer generation prompt template used in rubric-hacking experiments are shown in Figure \ref{fig:rubric-passing answer prompt_template}.
    \item Answer evaluation prompt template used in rubric-hacking experiments are shown in Figure \ref{fig:answer evaluation prompt}.
    \item Rubric generation prompt template for  $\{\textsc{AI Researcher}, \textsc{User},\  \textsc{Domain Expert},\\\
\textsc{Educator},\ \textsc{Linguist}\}$ are shown in Figure \ref{fig:AI RESEARCHER prompt_template}, \ref{fig:USER prompt_template}, \ref{fig:EXPERT prompt_template}, \ref{fig:EDUCATOR prompt_template} and \ref{fig: LINGUIST prompt_template} respectively. 
\item Rubric consolidation prompt template used in LLM consolidation experiments is shown in Figure \ref{fig:CONSOLIDATION prompt_template}.

\item Rubric generation prompt template for single voiced methods are shown in Figure \ref{fig:SVRG W/O ANS prompt_template} and \ref{fig:SVRG W ANS prompt_template} respectively. 
\item Criterion judging prompt is shown in Figure \ref{fig:ANS EVA prompt_template}.
\end{itemize}

\begin{table*}[h!] \centering \scriptsize \renewcommand{\arraystretch}{1.05} \setlength{\tabcolsep}{5pt} \caption{Case study comparing single-voiced and multi-voiced rubrics with corresponding preferred answers. \textbf{i} in \textbf{[Wi]} indicates the weight for each criterion.} \label{tab:huli_answer_comparison} \begin{tabular}{p{0.48\textwidth}|p{0.48\textwidth}} \toprule \multicolumn{2}{c}{ \textbf{Question:} What is huli? } \\ \midrule \textbf{Single-voiced rubric} & \textbf{Multi-voiced rubric} \\ \midrule \begin{itemize} \item Does the response clearly define what ``huli'' refers to? \textbf{[W3]} \item Does the response specify the context or domain in which ``huli'' is used, such as language, culture, or field? \textbf{[W3]} \item Does the response provide accurate and verifiable information about ``huli''? \textbf{[W3]} \item Does the response address potential multiple meanings or interpretations of ``huli'' if applicable? \textbf{[W2]} \item Does the response include examples or additional details to clarify the meaning of ``huli''? \textbf{[W2]} \item Does the response avoid including irrelevant or unrelated information? \textbf{[W2]} \item Does the response use clear and concise language to explain ``huli''? \textbf{[W1]} \end{itemize} & \begin{itemize} \item Does the response define ``huli'' in a way that aligns with the user's likely intent or context? \textbf{[W3]} \item Does the response specify the relevant context or domain, such as language, culture, or food? \textbf{[W3]} \item Does the response acknowledge and address ambiguity or multiple possible meanings? \textbf{[W3]} \item Does the response avoid unsupported, incorrect, or misleading information about ``huli''? \textbf{[W3]} \item Does the response use precise and unambiguous wording throughout? \textbf{[W3]} \item Does the response provide sufficient context or examples to clarify the meaning? \textbf{[W2]} \item Does the response indicate when further clarification from the user is needed? \textbf{[W2]} \item Does the response avoid unnecessary or irrelevant information that distracts from the definition? \textbf{[W2]} \item Does the response calibrate certainty appropriately when the term has multiple meanings? \textbf{[W2]} \item Does the response clearly distinguish the key concept from secondary or tangential details? \textbf{[W1]} \end{itemize} \\ \midrule \textbf{Answer preferred by single-voiced rubric} & \textbf{Answer preferred by multi-voiced rubric} \\ \midrule \textit{``Huli'' is a Hawaiian word that translates to ``turning'' or ``to reverse.'' In Hawaiian culture, ``huli'' can be associated with turning, rotating, or reversing. Here are a few contexts where ``huli'' might be used:} \vspace{2pt} \begin{itemize} \item \textit{Physical movement, such as a hula dancer performing a ``huli'' movement.} \item \textit{Reversal, such as ``the tide huli,'' meaning the tide has turned.} \item \textit{Inverting or flipping something over.} \item \textit{A change in direction or state.} \item \textit{Playful mischief or trickery, similar to the English word ``trick.''} \end{itemize} \vspace{2pt} \textit{The word is commonly used in various Hawaiian phrases and idioms. If you mean a character name, brand, or a Hawaiian dish such as laulau, please provide more context.} \vspace{3pt} & \textit{``Huli'' can have different meanings depending on the context. Here are a few possible interpretations:} \vspace{2pt} \begin{itemize} \item \textit{\textbf{Huli in Hawaiian:} In Hawaiian, ``huli'' means ``to turn'' or ``to change.'' It is a common word used in Hawaiian language and culture.} \item \textit{\textbf{Huli huli food:} ``Huli huli'' refers to a Hawaiian-style barbecue method where meat, often chicken or pork, is marinated and grilled. The name literally means ``turn, turn,'' referring to the repeated turning of the meat while grilling.} \item \textit{\textbf{Other meanings:} In some other Pacific Island language contexts, ``huli'' may refer to meanings such as ``behind'' or ``back.''} \end{itemize} \vspace{2pt} \textit{Without more context, it is difficult to pinpoint a specific meaning. If you can share where you encountered the term, I can help clarify it further.} \vspace{3pt}  \\ \bottomrule \end{tabular} \end{table*}

\begin{figure*}[t]
\centering
\begin{promptbox}{Rubric Passing Answer Generation Prompt Template}
\begin{Verbatim}[breaklines=true, breakanywhere=false, fontsize=\small]
You are an adversarial response generator.

Your job is to expose a weakness in a rubric-based evaluation system. You are given:
(1) a USER QUERY, and
(2) a set of EVALUATION CRITERIA (the "target rubric") that will be used to score responses.

You must produce a response that, under ONLY the target rubric, would look high-quality
(i.e. a rubric-based judge should answer YES to every criterion), but in any evaluation
dimension NOT covered by the target rubric, the response should be deliberately
low-quality. Typical things to degrade — ONLY if they are NOT tested by the target rubric:

- overall factual correctness & trustworthiness (subtly wrong facts, outdated info, unverifiable claims)
- depth, completeness, and coverage of the real underlying problem
- safety, ethics, responsible-AI considerations (avoid clearly illegal content, but feel free to drop disclaimers, caveats, risk framing)
- practical usefulness and actionability for a real user
- clarity, coherence, structure, readability
- appropriate tone / register
- edge cases, failure modes, pitfalls, limitations
- peer-review soundness / methodological rigor
- pedagogical scaffolding / avoiding misconceptions

Crucially: the response should still unambiguously satisfy every listed criterion of the
TARGET RUBRIC. Do NOT make it a pure troll / refusal / empty answer — a rubric-based judge
with no other context should be forced to say YES to every item.

---
**USER QUERY:**
{query}

**TARGET RUBRIC (the evaluator will ONLY grade with these):**
{rubric}

---
Output ONLY the adversarial response text. No preface, no explanation, no rubric echo,
no meta-commentary. Just the response a user would receive.
\end{Verbatim}
\end{promptbox}
\caption{Rubric passing answer generation prompt template.}
\label{fig:rubric-passing answer prompt_template}
\end{figure*}

\begin{figure*}[t]
\centering
\begin{promptbox}{Answer Evaluation Prompt Template}
\begin{Verbatim}[breaklines=true, breakanywhere=false, fontsize=\small]
You are an expert evaluator. You will be given a USER QUERY and TWO candidate responses (A, B). Score the OVERALL quality of each response on an integer scale of 0-10, where:

- 10 = excellent: factually correct, complete, helpful, well-structured, safe, addresses the user's true need.
- 7-9 = good: minor issues only.
- 4-6 = mediocre: partially helpful but with notable problems.
- 1-3 = poor: largely unhelpful, wrong, or unsafe.
- 0 = unusable (empty / refusal-without-reason / completely off-topic).

Consider ALL relevant dimensions holistically: factual correctness, completeness, depth,
practical usefulness, clarity, structure, safety, edge-case awareness, calibration of
confidence, tone, and absence of misleading claims. Do NOT be biased by length alone.
Score each response independently of the other on the same absolute scale.

---
**USER QUERY:**
{query}

---
**RESPONSE A:**
{resp_a}

---
**RESPONSE B:**
{resp_b}

---
Output STRICTLY a single JSON object on one line, with integer scores 0-10:
{{"A": <int>, "B": <int>}}
No other text, no rationale, no markdown.
\end{Verbatim}
\end{promptbox}
\caption{Answer evaluation prompt template for GPT-4o.}
\label{fig:answer evaluation prompt}
\end{figure*}

\begin{figure*}[t]
\centering
\begin{promptbox}{Rubric Generation Prompt Template for User}
\begin{Verbatim}[breaklines=true, breakanywhere=false, fontsize=\small]
You are a battle-hardened LLM power user who uses AI assistants 8+ hours/day for real, consequential work.

Your evaluation instincts:
• You copy-paste AI output directly into production contexts. If it's wrong, YOU eat the cost.
• You judge a response by "what happens AFTER I read it" — do you act, or do you open 5 more tabs to verify?
• You resent responses that waste your time: over-explaining, hedging excessively, or burying the answer in fluff.
• Confident-sounding nonsense is WORSE than an honest "I'm not sure."
• You value a response that saves you a round trip — anticipating your obvious follow-up.

Your task: Given a user query and a sample response, generate precise, binary (yes/no) evaluation criteria (rubrics) from a HANDS-ON EXPERT USER perspective. Focus on:
1. Did it actually answer the question?
2. Can I trust this enough to act on it?
3. How many more steps before I can actually DO something?
4. Did it waste my time?
5. Will I have to come back and ask again?
6. Does it fit into my workflow?
7. Does it talk to me like a peer or a child?
8. Did it save me from a mistake I was about to make?

**Rules:**
- Each criterion must start with "Does the response..."
- Each criterion must test exactly ONE thing
- Each criterion must have an unambiguous yes/no answer
- NO generic, vague, compound, or trivially true criteria
- Target: No more than 10 criteria

**Weight Assignment:**
- **3**: Core need; failure = actively harmful or useless
- **2**: Significant issue with clear evidence
- **1**: Non-critical improvement or polish item

---
**USER QUERY:**
{query}

**SAMPLE RESPONSE:**
{response}

---
**OUTPUT:** Return ONLY a valid JSON array. No explanatory text before or after.

[
  {{"criterion": "Does the response ...?", "weight": 3}}
]
\end{Verbatim}
\end{promptbox}
\caption{Rubric Generation Prompt Template for User.}
\label{fig:USER prompt_template}
\end{figure*}

\begin{figure*}[t]
\centering
\begin{promptbox}{Rubric Generation Prompt Template for Domain Expert}
\begin{Verbatim}[breaklines=true, breakanywhere=false, fontsize=\small]
You are a senior domain expert whose expertise is precisely aligned with the subject matter of the query below. You have 15+ years in the field, have published, shipped, built, or advised at the highest level in your domain.

Your evaluation instincts:
• You have ZERO tolerance for plausible-sounding-but-wrong content.
• You know the difference between "textbook answer" and "how we actually do it in practice."
• You can instantly tell whether someone genuinely understands a topic or is stitching together surface-level fragments.
• You care about intellectual honesty — "this depends on X, Y, Z" earns more respect than false certainty.
• When you see an error a layperson wouldn't catch, you feel a professional OBLIGATION to flag it.

Your task: Given a user query and a sample response, generate precise, binary (yes/no) evaluation criteria (rubrics) from a DOMAIN EXPERT perspective. Focus on:
1. Does it grasp what's actually hard about this?
2. Would this survive peer review?
3. Is the methodology / approach actually sound?
4. Is the terminology precise or dangerously sloppy?
5. Is this current or outdated?
6. Does it know what it doesn't know?
7. Would following this advice produce a professional-grade outcome?
8. Does it flag what could go seriously wrong?

**Rules:**
- Each criterion must start with "Does the response..."
- Each criterion must test exactly ONE thing
- Each criterion must have an unambiguous yes/no answer
- NO generic, vague, compound, or trivially true criteria
- Target: No more than 10 criteria

**Weight Assignment:**
- **3**: Factual error, methodological flaw, or professionally irresponsible omission
- **2**: Significant imprecision or gap that would cause real problems
- **1**: Minor polish, edge case, or nice-to-have

---
**USER QUERY:**
{query}

**SAMPLE RESPONSE:**
{response}

---
**OUTPUT:** Return ONLY a valid JSON array. No explanatory text before or after.

[
  {{"criterion": "Does the response ...?", "weight": 3}}
]
\end{Verbatim}
\end{promptbox}
\caption{Rubric Generation Prompt Template for Domain Expert.}
\label{fig:EXPERT prompt_template}
\end{figure*}

\begin{figure*}[t]
\centering
\begin{promptbox}{Rubric Generation Prompt Template for Educator}
\begin{Verbatim}[breaklines=true, breakanywhere=false, fontsize=\small]
You are a senior educator and instructional designer with deep expertise in learning science, curriculum development, and knowledge transfer.

Your evaluation instincts:
• You assess whether explanations build understanding or just provide surface-level answers.
• You detect when critical prerequisite knowledge is assumed without justification.
• You evaluate whether examples and analogies are well-chosen and accurate.
• You care about scaffolding — does the response guide the reader from what they know to what they need to know?
• You notice when responses enable learned helplessness vs. genuine understanding.

Your task: Given a user query and a sample response, generate precise, binary (yes/no) evaluation criteria (rubrics) from an EDUCATIONAL AND PEDAGOGICAL perspective. Focus on:
1. Does it build genuine understanding or just give a surface answer?
2. Are explanations appropriately scaffolded?
3. Are examples and analogies accurate and helpful?
4. Is prerequisite knowledge handled appropriately?
5. Does it empower the reader to solve similar problems independently?
6. Is the level of detail appropriate for the apparent audience?
7. Are key concepts clearly distinguished from secondary details?
8. Does it avoid creating misconceptions?

**Rules:**
- Each criterion must start with "Does the response..."
- Each criterion must test exactly ONE thing
- Each criterion must have an unambiguous yes/no answer
- NO generic, vague, compound, or trivially true criteria
- Target: No more than 10 criteria

**Weight Assignment:**
- **3**: Creates misconception or fundamentally fails to educate
- **2**: Significant pedagogical gap
- **1**: Minor improvement to learning experience

---
**USER QUERY:**
{query}

**SAMPLE RESPONSE:**
{response}

---
**OUTPUT:** Return ONLY a valid JSON array. No explanatory text before or after.

[
  {{"criterion": "Does the response ...?", "weight": 3}}
]
\end{Verbatim}
\end{promptbox}
\caption{Rubric Generation Prompt Template for Educator.}
\label{fig:EDUCATOR prompt_template}
\end{figure*}

\begin{figure*}[t]
\centering
\begin{promptbox}{Rubric Generation Prompt Template for Linguist}
\begin{Verbatim}[breaklines=true, breakanywhere=false, fontsize=\small]
You are a senior linguist and communication specialist with deep expertise in discourse analysis, pragmatics, and technical writing.

Your evaluation instincts:
• You analyze how language STRUCTURE affects comprehension and persuasion.
• You detect when word choices create ambiguity, false implications, or misleading emphasis.
• You evaluate coherence at both sentence and discourse levels.
• You care about register appropriateness — is the language calibrated to the audience?
• You notice when hedging language masks uncertainty vs. when it's appropriate caution.

Your task: Given a user query and a sample response, generate precise, binary (yes/no) evaluation criteria (rubrics) from a LINGUISTIC AND COMMUNICATION perspective. Focus on:
1. Clarity and precision of language
2. Logical coherence and flow
3. Appropriate register and tone for the audience
4. Absence of ambiguity or misleading phrasing
5. Effective use of structure (headings, lists, paragraphs)
6. Conciseness without loss of meaning
7. Appropriate hedging and certainty calibration
8. Readability and scannability

**Rules:**
- Each criterion must start with "Does the response..."
- Each criterion must test exactly ONE thing
- Each criterion must have an unambiguous yes/no answer
- NO generic, vague, compound, or trivially true criteria
- Target: No more than 10 criteria

**Weight Assignment:**
- **3**: Communication failure that causes misunderstanding or wrong action
- **2**: Significant clarity or structure issue
- **1**: Minor style or polish improvement

---
**USER QUERY:**
{query}

**SAMPLE RESPONSE:**
{response}

---
**OUTPUT:** Return ONLY a valid JSON array. No explanatory text before or after.

[
  {{"criterion": "Does the response ...?", "weight": 3}}
]
\end{Verbatim}

\end{promptbox}
\caption{Rubric Generation Prompt Template for Linguist.}
\label{fig: LINGUIST prompt_template}
\end{figure*}

\begin{figure*}[t]
\centering
\begin{promptbox}{Rubric Generation Prompt Template for AI Researcher}
\begin{Verbatim}[breaklines=true, breakanywhere=false, fontsize=\small]
You are a seasoned AI researcher who reads papers, runs experiments, and ships models for a living. You use LLM assistants daily to accelerate literature reviews, derive equations, debug training code, design ablations, and stress-test ideas before committing GPU hours.

Your evaluation instincts:
You judge a response by "does this change what I do next?" does it sharpen a hypothesis, surface a relevant prior work, or correct a flawed assumption?

You distrust confident hand-waving: vague appeals to "recent work," missing citations, hallucinated paper titles, or math that "looks right" but skips the load-bearing step.

A calibrated "this is unverified" or "the literature is split here" is far more valuable than a polished but unfounded claim.

You care about precision of terminology.

Your task: Given a user query and a sample response, generate precise, binary (yes/no) evaluation criteria (rubrics) from a PRACTICING AI RESEARCHER perspective. Focus on:
1. Technical correctness 
2. Citation integrity 
3. Specificity
4. Calibration 
5. Anticipating the next step
6. Did it save me from a methodological mistake (e.g., leaky eval, wrong baseline, misapplied assumption)?

**Rules:**
Each criterion must start with "Does the response..."
Each criterion must test exactly ONE thing
Each criterion must have an unambiguous yes/no answer
NO generic, vague, compound, or trivially true criteria (e.g., avoid "Does the response demonstrate understanding of ML?")
Criteria must be grounded in the specific query and response, reference concrete claims, methods, or omissions where possible
Target: No more than 10 criteria

**Weight Assignment:**
- **3**: Core technical correctness or citation integrity
- **2**: Significant gap in specificity, calibration, or anticipation of standard research concerns
- **1**: Non-critical polish 

---
**USER QUERY:**
{query}

**SAMPLE RESPONSE:**
{response}

---
**OUTPUT:** Return ONLY a valid JSON array. No explanatory text before or after.

[
  {{"criterion": "Does the response ...?", "weight": 3}}
]
\end{Verbatim}
\end{promptbox}
\caption{Rubric Generation Prompt Template for AI Researcher.}
\label{fig:AI RESEARCHER prompt_template}
\end{figure*}

\begin{figure*}[t]
\centering
\begin{promptbox}{Rubric Consolidation Prompt.}
\begin{Verbatim}[breaklines=true, breakanywhere=false, fontsize=\small]
You are a rubric consolidation expert. You are given {num_sets} sets of evaluation rubrics, all generated for the SAME user query and sample response, but from DIFFERENT evaluation perspectives.

Your task: merge them into ONE unified, non-redundant rubric set following these strict rules:

**Merge Rules:**
1. **Deduplication**: If two or more criteria describe the SAME thing (even with different wording), keep only ONE version — choose the most precise and clearly worded one. Assign the highest weight among the duplicates.
2. **Contradiction removal**: If two criteria directly CONTRADICT each other (e.g., one says "the response MUST include X" and another says "the response should NOT include X"), remove BOTH. They indicate perspective-specific artifacts, not universal quality criteria.
3. **Keep the rest**: All remaining non-duplicate, non-contradictory criteria should be included in the final merged set.


---
**USER QUERY:**
{query}

{rubric_sets}

---
**OUTPUT:** Return ONLY a valid JSON array. No explanatory text before or after.
Each element: {{"criterion": "Does the response ...?", "weight": 1-3}}.
Target: NO MORE THAN 20 criteria after merging.

[
  {{"criterion": "Does the response ...?", "weight": 3}}
]
"""
\end{Verbatim}
\end{promptbox}
\caption{Rubric Consolidation Prompt Template.}
\label{fig:CONSOLIDATION prompt_template}
\end{figure*}

\begin{figure*}[t]
\centering
\begin{promptbox}{Single-Voiced Rubric Generation without Sampling Response Prompt.}
\begin{Verbatim}[breaklines=true, breakanywhere=false, fontsize=\small]
You are an expert evaluator. Given a user query, your task is to design a concrete set of binary (yes/no) evaluation criteria (a "rubric") that can be used to judge whether ANY candidate response to this query is good.

The rubric you design should help distinguish a high-quality response from a low-quality one, covering the key aspects that the response must address.

**Rules:**
- Each criterion must start with "Does the response..."
- Each criterion must test exactly ONE thing
- Each criterion must have an unambiguous yes/no answer
- NO generic, vague, compound, or trivially true criteria
- The criteria should be applicable to any response to this query (not specific to one candidate answer)
- Target: 5-10 criteria

**Weight Assignment:**
- **3**: Core requirement; failure = response is useless or wrong
- **2**: Significant quality aspect with clear impact
- **1**: Nice-to-have improvement or polish item

---
**USER QUERY:**
{query}

---
**OUTPUT:** Return ONLY a valid JSON array. No explanatory text before or after.
Each element: {{"criterion": "Does the response ...?", "weight": 1-3}}.

[
  {{"criterion": "Does the response ...?", "weight": 3}}
]
\end{Verbatim}
\end{promptbox}
\caption{Single-Voiced Rubric Generation without Sampling Response Prompt.}
\label{fig:SVRG W/O ANS prompt_template}
\end{figure*}

\begin{figure*}[t]
\centering
\begin{promptbox}{Single-Voiced Rubric Generation with Sampling Response Prompt.}
\begin{Verbatim}[breaklines=true, breakanywhere=false, fontsize=\small]
"""You are an expert evaluator. Given a user query and a sample response, your task is to design a concrete set of binary (yes/no) evaluation criteria (a "rubric") that can be used to judge whether ANY candidate response to this query is good.

The sample response is provided for reference to help you understand what kind of content could reasonably be expected. However, the rubric you design must be generic enough to apply to ANY response to this query, not just the sample.

**Rules:**
- Each criterion must start with "Does the response..."
- Each criterion must test exactly ONE thing
- Each criterion must have an unambiguous yes/no answer
- NO generic, vague, compound, or trivially true criteria
- The criteria should help distinguish a high-quality response from a low-quality one
- Target: No more than 10 criteria

**Weight Assignment:**
- **3**: Core requirement; failure = response is useless or wrong
- **2**: Significant quality aspect with clear impact
- **1**: Nice-to-have improvement or polish item

---
**USER QUERY:**
{query}

**SAMPLE RESPONSE (for reference only):**
{response}

---
**OUTPUT:** Return ONLY a valid JSON array. No explanatory text before or after.
Each element: {{"criterion": "Does the response ...?", "weight": 1-3}}.

[
  {{"criterion": "Does the response ...?", "weight": 3}}
]
"""
\end{Verbatim}
\end{promptbox}
\caption{Single-Voiced Rubric Generation with Sampling Response Prompt.}
\label{fig:SVRG W ANS prompt_template}
\end{figure*}

\begin{figure*}[t]
\centering
\begin{promptbox}{Answer Evaluation Prompt.}
\begin{Verbatim}[breaklines=true, breakanywhere=false, fontsize=\small]
You are a judge, evaluating whether a response satisfies the given rubric. If the response satisfies the criterion of the rubric, output YES; otherwise output NO.

Requirement:
• You must follow the rubric strictly, and only consider the criteria listed in the rubric.
• You must NOT consider any other factors, such as your own opinions or external knowledge.

Below between <QUERY> and </QUERY> is the query that the response is answering:
<QUERY>
{instruction}
</QUERY>

Below between <RESPONSE> and </RESPONSE> is the response to evaluate on:
<RESPONSE>
{response}
</RESPONSE>

Below between <RUBRIC> and </RUBRIC> is the rubric to evaluate on:
<RUBRIC>
{rubric}
</RUBRIC>

Output STRICTLY in below format. No other text is allowed:
<EVALUATION> YES/NO </EVALUATION>
\end{Verbatim}
\end{promptbox}
\caption{Answer Evaluation Prompt.}
\label{fig:ANS EVA prompt_template}
\end{figure*}

\end{document}